\documentclass[12pt]{article}
\usepackage{geometry}
\usepackage{amsthm}
\usepackage{amsmath}
\usepackage{amssymb}
\usepackage{dsfont}
\usepackage{tikz,pgfplots}
\usepackage[colorlinks=true,breaklinks=true,bookmarks=false,urlcolor=blue,%
  citecolor=blue,linkcolor=blue,bookmarksopen=false,draft=false]{hyperref}
\usepackage{algorithm}
\usepackage{algpseudocode}
\usepackage{subfigure}

\newcommand{\E}{\mathbb{E}}
\newcommand{\R}{\mathbb{R}}
\renewcommand{\P}{\mathbb{P}}

\newcommand{\Rdx}{\mathbb{R}^{d_x}}
\newcommand{\Rdy}{\mathbb{R}^{d_y}}
\newcommand{\Rdw}{\mathbb{R}^{d_w}}
\newcommand{\Xcal}{{\cal X}}
\newcommand{\Ycal}{{\cal Y}}
\newcommand{\Wcal}{{\cal W}}
\newcommand{\fhat}{\hat{f}}
\newcommand{\what}{\hat{w}}
\newcommand{\Ocal}{{\cal {O}}}
\usetikzlibrary{patterns}
\usetikzlibrary{arrows}

\begin{document}
\title{Optimization Methods for Supervised Machine Learning: From Linear Models to Deep Learning}

  \author{Frank E.~Curtis\thanks{Lehigh University, Department of ISE, \href{mailto:frank.e.curtis@lehigh.edu}{frank.e.curtis@lehigh.edu}}
  \and
          Katya Scheinberg\thanks{Lehigh University, Department of ISE, \href{mailto:katyas@lehigh.edu}{katyas@lehigh.edu}} 
}

\maketitle  

\abstract{%
The goal of this tutorial is to introduce key models, algorithms, and open questions related to the use of optimization methods for solving problems arising in machine learning.  It is written with an INFORMS audience in mind, specifically those readers who are familiar with the basics of optimization algorithms, but less familiar with machine learning.  We begin by deriving a formulation of a supervised learning problem and show how it leads to various optimization problems, depending on the context and underlying assumptions.  We then discuss some of the distinctive features of these optimization problems, focusing on the examples of logistic regression and the training of deep neural networks.  The latter half of the tutorial focuses on optimization algorithms, first for convex logistic regression, for which we discuss the use of first-order methods, the stochastic gradient method, variance reducing stochastic methods, and second-order methods.  Finally, we discuss how these approaches can be employed to the training of deep neural networks, emphasizing the difficulties that arise from the complex, nonconvex structure of these models.
} 


%

\section{Introduction}

The past two decades have witnessed the almost unprecedented rise of an intriguing algorithmic field: machine learning (ML).  With roots in statistics and computer science, ML has a mathematical optimization engine at its core.  In fact, these days, ML and other data-driven disciplines have influenced much of the latest theoretical and practical advances in the optimization research community.  Still, despite these connections, many barriers remain between the statistics, computer science, and optimization communities working on ML and related topics.  These barriers---of differing terminology, goals, and avenues for collaboration---have led to duplications of efforts and continue to inhibit effective exchanges of ideas.

The aim of this tutorial is to present an overview of some of the key issues and research questions that relate to optimization within the field of ML.  With the Operations Research (OR) community in mind, we assume that the reader is familiar with basic optimization methodologies, but will introduce terminology and concepts used in the broader ML community in a manner that we hope will facilitate communication between OR experts and those from other contributing fields.  To aid in this pursuit, we provide in Table~\ref{tab.glossary} a small glossary of the most important terms that will be introduced and used throughout this tutorial.

\begin{table}[ht]
  \centering
  \caption{Glossary of terms in supervised machine learning where one aims to understand a relationship between each input $x$ from a space ${\cal X}$ and its corresponding output $y$ in a space~${\cal Y}$.}
  \label{tab.glossary}
  \footnotesize
  \begin{tabular}{ccll}
    \hline
    Term & Notation & \multicolumn{1}{c}{Definition} & \multicolumn{1}{c}{a.k.a.} \\
    \hline
    input & $x$ & element from input space ${\cal X}$ & feature (vector) \\
    output & $y$ & element from output space ${\cal Y}$ & label (vector) \\
    sample set & $\{(x_i,y_i)\}_{i=1}^n$ & pairs from input $\times$ output space & \\
    testing set &  & other set of pairs from input $\times$ output space \\
    prediction function & $p$ & function such that $p(x)$ predicts $y$ & predictor, \\ &&& classifier \\
    parameter vector & $w$ & parameterization vector for prediction function, & weights \\
    && i.e., $p \equiv p(w,\cdot)$ with $w$ in a space ${\cal W}$ \\
    loss function & $\ell$ & function for assigning penalty when $p(x)$ \\ && does not predict the correct label for $x$ \\
    training/testing loss &  & average loss evaluated over sample/testing set \\
    training/testing error &  & percentage of mislabeled elements \\ && in the sample/testing set \\
    stepsize sequence & $\{\alpha_k\}$ & multipliers for steps in optimization algorithm & learning rate \\
    \hline
  \end{tabular}
\end{table}

\subsection{Motivating illustration}\label{sec.illustration}

The idea of machine learning arises with the fundamental question of whether machines (i.e., computers) can ``think'' like humans.  At a more tangible level, this leads to questions such as whether, given a particular input, a machine can produce a reasonable/rational output response, e.g., for the purpose of making a prediction or decision.  Let us begin with a simple illustration of how this might be done before introducing the idea of a \emph{learning problem} more formally in the next subsection.

Suppose that a company aims to predict whether \emph{Product A} is going to be profitable (yes or no) in an upcoming quarter.  A human expert might attempt to make such a prediction by considering various factors that can be found in historical data, say about \emph{Product A}, related products, and/or other factors.  For simplicity of illustration, suppose that one considers two factors, also known as \emph{features}: the demand for \emph{Product A} and another factor, call it \emph{Factor~X}, both projected for the upcoming quarter.  Certainly, one might expect that projected high demand might suggest high potential profitability in the upcoming quarter, but compounded with the influence of \emph{Factor X} the outcome might be less obvious. For example, \emph{Factor X} may reflect the cost of production or delivery of \emph{Product A}, which might depend on the costs of raw materials or the set-up (take-down) costs of ramping up (reducing) production levels compared to previous quarters.  Overall, the influence of \emph{Factor~X} could be complex and nonlinear.

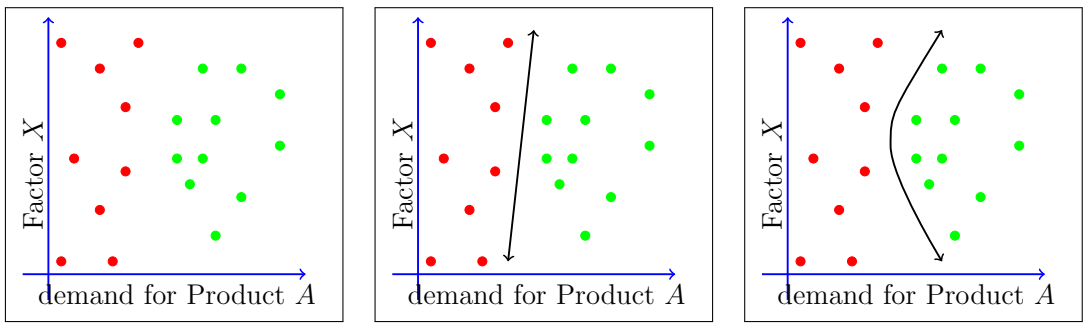
\begin{figure}[ht]
  \centering
  \fbox{\scalebox{0.9}{\begin{tikzpicture}[scale=1.9]
  \coordinate (orig) at ( 0.0, 0.0);
  \coordinate (xmax) at ( 2.0, 0.0);
  \coordinate (ymax) at ( 0.0, 2.0);
  \coordinate (xmin) at (-0.2, 0.0);
  \coordinate (ymin) at ( 0.0,-0.2);
  \draw[blue,thick,->] (xmin) -- (xmax);
  \draw[blue,thick,->] (ymin) -- (ymax);
  \coordinate[label=below:demand for Product $A$] (xmid) at ( 1.0, 0.0);
  \coordinate[label={[rotate=90]left:Factor $X$}]  (ymid) at (-0.12, 1.3);
  \draw[red,fill=red] (0.1,0.1) circle (1pt);
  \draw[red,fill=red] (0.1,1.8) circle (1pt);
  \draw[red,fill=red] (0.2,0.9) circle (1pt);
  \draw[red,fill=red] (0.4,1.6) circle (1pt);
  \draw[red,fill=red] (0.4,0.5) circle (1pt);
  \draw[red,fill=red] (0.5,0.1) circle (1pt);
  \draw[red,fill=red] (0.6,0.8) circle (1pt);
  \draw[red,fill=red] (0.6,1.3) circle (1pt);
  \draw[red,fill=red] (0.7,1.8) circle (1pt);
  \draw[green,fill=green] (1.0,0.9) circle (1pt);
  \draw[green,fill=green] (1.0,1.2) circle (1pt);
  \draw[green,fill=green] (1.1,0.7) circle (1pt);
  \draw[green,fill=green] (1.2,0.9) circle (1pt);
  \draw[green,fill=green] (1.2,1.6) circle (1pt);
  \draw[green,fill=green] (1.3,0.3) circle (1pt);
  \draw[green,fill=green] (1.3,1.2) circle (1pt);
  \draw[green,fill=green] (1.5,0.6) circle (1pt);
  \draw[green,fill=green] (1.5,1.6) circle (1pt);
  \draw[green,fill=green] (1.8,1.0) circle (1pt);
  \draw[green,fill=green] (1.8,1.4) circle (1pt);
\end{tikzpicture}}}\quad 
  \fbox{\scalebox{0.9}{\begin{tikzpicture}[scale=1.9]
  \coordinate (orig) at ( 0.0, 0.0);
  \coordinate (xmax) at ( 2.0, 0.0);
  \coordinate (ymax) at ( 0.0, 2.0);
  \coordinate (xmin) at (-0.2, 0.0);
  \coordinate (ymin) at ( 0.0,-0.2);
  \draw[blue,thick,->] (xmin) -- (xmax);
  \draw[blue,thick,->] (ymin) -- (ymax);
  \coordinate[label=below:demand for Product $A$] (xmid) at ( 1.0, 0.0);
  \coordinate[label={[rotate=90]left:Factor $X$}]  (ymid) at (-0.12, 1.3);
  \draw[thick,<->] (0.9,1.9) -- (0.7,0.1);
  \draw[red,fill=red] (0.1,0.1) circle (1pt);
  \draw[red,fill=red] (0.1,1.8) circle (1pt);
  \draw[red,fill=red] (0.2,0.9) circle (1pt);
  \draw[red,fill=red] (0.4,1.6) circle (1pt);
  \draw[red,fill=red] (0.4,0.5) circle (1pt);
  \draw[red,fill=red] (0.5,0.1) circle (1pt);
  \draw[red,fill=red] (0.6,0.8) circle (1pt);
  \draw[red,fill=red] (0.6,1.3) circle (1pt);
  \draw[red,fill=red] (0.7,1.8) circle (1pt);
  \draw[green,fill=green] (1.0,0.9) circle (1pt);
  \draw[green,fill=green] (1.0,1.2) circle (1pt);
  \draw[green,fill=green] (1.1,0.7) circle (1pt);
  \draw[green,fill=green] (1.2,0.9) circle (1pt);
  \draw[green,fill=green] (1.2,1.6) circle (1pt);
  \draw[green,fill=green] (1.3,0.3) circle (1pt);
  \draw[green,fill=green] (1.3,1.2) circle (1pt);
  \draw[green,fill=green] (1.5,0.6) circle (1pt);
  \draw[green,fill=green] (1.5,1.6) circle (1pt);
  \draw[green,fill=green] (1.8,1.0) circle (1pt);
  \draw[green,fill=green] (1.8,1.4) circle (1pt);
\end{tikzpicture}}}\quad 
  \fbox{\scalebox{0.9}{\begin{tikzpicture}[scale=1.9]
  \coordinate (orig) at ( 0.0, 0.0);
  \coordinate (xmax) at ( 2.0, 0.0);
  \coordinate (ymax) at ( 0.0, 2.0);
  \coordinate (xmin) at (-0.2, 0.0);
  \coordinate (ymin) at ( 0.0,-0.2);
  \draw[blue,thick,->] (xmin) -- (xmax);
  \draw[blue,thick,->] (ymin) -- (ymax);
  \coordinate[label=below:demand for Product $A$] (xmid) at ( 1.0, 0.0);
  \coordinate[label={[rotate=90]left:Factor $X$}]  (ymid) at (-0.12, 1.3);
  \draw[thick,<->] plot [smooth] coordinates {(1.2,1.9) (0.8,1.0) (1.2,0.1)};
  \draw[red,fill=red] (0.1,0.1) circle (1pt);
  \draw[red,fill=red] (0.1,1.8) circle (1pt);
  \draw[red,fill=red] (0.2,0.9) circle (1pt);
  \draw[red,fill=red] (0.4,1.6) circle (1pt);
  \draw[red,fill=red] (0.4,0.5) circle (1pt);
  \draw[red,fill=red] (0.5,0.1) circle (1pt);
  \draw[red,fill=red] (0.6,0.8) circle (1pt);
  \draw[red,fill=red] (0.6,1.3) circle (1pt);
  \draw[red,fill=red] (0.7,1.8) circle (1pt);
  \draw[green,fill=green] (1.0,0.9) circle (1pt);
  \draw[green,fill=green] (1.0,1.2) circle (1pt);
  \draw[green,fill=green] (1.1,0.7) circle (1pt);
  \draw[green,fill=green] (1.2,0.9) circle (1pt);
  \draw[green,fill=green] (1.2,1.6) circle (1pt);
  \draw[green,fill=green] (1.3,0.3) circle (1pt);
  \draw[green,fill=green] (1.3,1.2) circle (1pt);
  \draw[green,fill=green] (1.5,0.6) circle (1pt);
  \draw[green,fill=green] (1.5,1.6) circle (1pt);
  \draw[green,fill=green] (1.8,1.0) circle (1pt);
  \draw[green,fill=green] (1.8,1.4) circle (1pt);
\end{tikzpicture}}}
  \caption{On the left, pairs of historical data of predicted demand of \emph{Product A} and \emph{Factor~X}, where green indicates a resulting profitable quarter and red indicates an unprofitable one.  In the center and right, the data points separated by a linear or nonlinear classifier, respectively.}
  \label{fig.product_A}
\end{figure}

Looking at historical data over multiple consecutive quarters, suppose that the data points in Figure~\ref{fig.product_A} show the pairs of demand for \emph{Product~A} and value of \emph{Factor~X} that have been observed.  The points in green indicate pairs corresponding to quarters in which \emph{Product~A} was profitable while those in red indicate unprofitable quarters.  Using this data, one might aim to \emph{learn} how to take the inputs of projected demand and \emph{Factor~X} and predict whether or not the product will be profitable.  For example, this could be achieved by  learning a dividing line  or a dividing curve between green and red dots, as illustrated in the center and rightmost plots in Figure~\ref{fig.product_A}, respectively.  The idea is that, if a good prediction tool (i.e., dividing line/curve) is determined, then one could take a new pair of inputs and accurately predict whether \emph{Product~A} will be profitable in the upcoming quarter (in this example, by determining which side of the dividing line lies the new pair of inputs).

Of course, our illustration in Figure~\ref{fig.product_A} presents an idealized case in which dividing lines exist between data points corresponding to differing labels.  The situation is not always so clean.  We address this and other issues next in our more general discussion of learning.

\subsection{Learning problems and (surrogate) optimization problems}\label{sec.learning_problem}

Let us now define a generic learning problem more formally.  In particular, let us consider the problem of trying to determine an effective procedure for taking an input (i.e., \emph{feature} vector)~$x$ from an input space $\Xcal\subseteq \R^{d_x}$ and predicting what is its correct output (i.e., \emph{label} vector)~$y$ in an output space $\Ycal\subseteq \R^{d_y}$.  This can be cast as trying to learn a \emph{prediction} function, call it $p : \Xcal \to \Ycal$, that takes an input $x$ and produces $p(x)$ that one hopes can identify $y$.  Not knowing what might be future inputs of interest, one tries to determine~$p$ such that, over the distribution $P : \Xcal \times \Ycal \to \R$ of pairs in the input $\times$ output space $({\cal X},{\cal Y})$, one maximizes the probability of a correct prediction; i.e., the goal is to choose $p$ to maximize
\begin{equation}\label{eq.probability_correct}
  \int_{\Xcal \times \Ycal} \mathds{1}[p(x) \approx y] dP(x,y).
\end{equation}
Here, $\mathds{1}$ is the indicator function that takes the value 1 if its argument is true and 0 otherwise.  As for the notation ``$\approx$'', it could literally mean that the two vectors are equal (or close in some sense), that one is less than or equal to the other by some measure, or some other related meaning.  We discuss various such possibilities later on.


Clearly, for any given application, there is not necessarily one correct manner in which the specifics of the function $p$ in~\eqref{eq.probability_correct} should be chosen.  In addition, in its present form, it is far from tractable to maximize~\eqref{eq.probability_correct} directly.  In the remainder of this section, we discuss how, with various approximations and manipulations, one can go from the generic learning goal of maximizing~\eqref{eq.probability_correct} to a practical optimization problem that one can reasonably solve.

The first issue that one must address when aiming to maximize~\eqref{eq.probability_correct} is what class of prediction functions to consider.  If the class is too large and/or involves a diverse collection of complex functions, then finding that which maximizes~\eqref{eq.probability_correct} 
can be extremely difficult.   On the other hand, if the class is too small or only involves simple functions, then even the optimal $p$ within the class might yield poor predictions for various inputs.  (Recall the choice between trying to separate points based on a line or a more general type of curve in~\S\ref{sec.illustration}.)  Overall, there is a \emph{critical} balance that one should attempt to find when determining the class of prediction functions; see  \S\ref{sec:learningbounds} and \cite{BottCurtNoce16} for further discussion.

For our purposes, let us assume that some family of prediction functions is chosen that is parameterized by a real vector $w \in \Wcal = \Rdw$, i.e., $p \equiv p(w,\cdot)$.  For example, in a simple, yet very common setting, one considers the class of linear predictors where $p(w,x) = w_0+w_1^Tx$ with $w_0 \in \R$ and $w_1 \in \Rdx$, making the entire space of parameters $\Wcal = \R^{d_x+1}$.  Another example is the class of quadratic functions in $x$, where $p(w,x) = w_0 + w_1^Tx + w_2^T{\rm svec}(xx^T)$ with $w_0 \in \R$, $w_1 \in \Rdx$, and $w_2 \in \R^{d_x(d_x+1)/2}$ while ${\rm svec}(xx^T)$ denotes the $d_x(d_x+1)/2$-dimensional symmetric vectorization of the matrix $xx^T$,\footnote{The symmetric vectorization of $xx^T$ produces a vector whose elements correspond to the upper triangle of the outer product $xx^T$, namely terms of the form $[x_a]^2$ and $[x_a][x_b]$.} thus making ${\cal W}= {\bf R}^{(d_x+1)(d_x+2)/2}$. While this function is nonlinear in $x$---and hence can fit complex relationships between $x$ and $y$---it is linear in the parameters $w = (w_0,w_1, w_2)$.  Hence, this class can actually be viewed as a class of linear predictors for the modified feature space ${\cal X'}$, where for each $x \in {\cal X}$ one has $x'=(x,{\rm svec}(xx^T))$ in ${\cal X'}$.  It turns out that many, but not all, classes of prediction functions can be viewed as linear predictors in some modified space.  A notable exception of interest arises in the use of deep neural networks; see~\S\ref{sec.dnn}. 

Our learning problem can now be cast as trying to determine the parameters solving
\begin{equation}\label{eq:MLstoch}
  \max_{w\in\Wcal} \int_{\Xcal \times \Ycal} \mathds{1}[p(w,x) \approx y] dP(x,y).
\end{equation}
As a next step toward tractability, let us now discuss possible meanings for the expression $p(w,x) \approx y$.  Typically, the meaning depends on the type of labels involved.  For example:
\begin{itemize}
  \item In binary classification (``yes/no'' labeling) with $y \in \{+1,-1\}$, the expression $p(w,x) \approx y$ can represent $yp(w,x) > 0$.  In this manner, we say that the predicted value $p(w,x)$ identifies the correct output $y$ as long as the two values have the same sign.
  \item In regression with $y \in {\cal Y} \subseteq \Rdy$, the expression $p(w,x) \approx y$ might represent the fact that $\|y - p(w,x)\| \leq \delta$, where $\delta > 0$ is some prescribed accuracy threshold.
  \item In multi-class classification with $y \in \{0,1\}^{d_y}$ such that $\sum_{i=1}^{d_y} y_i=1$, the expression $p(w,x) \approx y$ might represent that $j(w,x) := \arg\max_{j\in\{1,\dots,d_y\}} p_j(w,x)$ is the index such that $y_{j(w,x)} = 1$.  In this manner, one can view the $j$th element of $p(w,x)/\|p(w,x)\|_1$ as some predicted probability that the true label of $x$ is $y_j$, and the label that will be predicted is the one with the largest predicted probability.
\end{itemize}

Let us continue our development toward a tractable problem by taking binary classification as a particular example. If we use $yp(w,x) > 0$ to represent
$p(w,x) \approx y$, then~\eqref{eq:MLstoch} becomes
\begin{equation}\label{eq:binary}
  \max_{w\in\Wcal} \int_{\Xcal \times \Ycal} \mathds{1}[yp(w,x) > 0] dP(x,y).
\end{equation}
This objective is very easy to interpret; it is the probability that the value $p(w,x)$ correctly predicts the sign of $y$. The problem can be rewritten
as 
\begin{equation}\label{eq:expected_risk01}
  \min_{w\in\Wcal}\ f(w),\ \ \text{where}\ \ f(w) := \int_{\Xcal \times \Ycal} \mathds{1}[yp(w,x)\leq 0] dP(x,y).
\end{equation}
The indicator $ \mathds{1}[yp(w,x)\leq 0]$ is known as the $01$-loss function.  It counts a unit loss if $p(w,x)$ incorrectly predicts the sign of $y$, and no loss otherwise.  This makes sense to do, but there are some drawbacks to using this loss function.  For one thing, it is discontinuous, making the goal of optimizing over $w$ potentially difficult.  In addition, and perhaps more importantly, this loss function does not quantify the {\em magnitude} of the error; e.g., for a given $w$, small perturbations in the data can cause large perturbations in $f(w)$, even though other (large) perturbations might not affect $f(w)$ at all.  This can lead to instability.  To overcome these issues, one can replace the $01$-loss by a similar, yet continuous (and perhaps smooth) surrogate loss function.  One way in which this can be done is through the idea of logistic regression, which can be described as follows.  First, imagine that the label is not deterministic.  Instead, given an input $x$, let $Y$ represent the random variable corresponding to the correct label for $x$.  The goal is to choose a parameter vector $w$ such that
\begin{equation}\label{eq.half}
  \begin{aligned}
  yp(w,x) > 0 &\iff \P(Y=y|x) > \tfrac12 \\ \text{while}\ \ yp(w,x) < 0 &\iff \P(Y=y|x) < \tfrac12.
  \end{aligned}
\end{equation}
Since $yp(w,x) \in \R$ while $\P(Y=y|x) \in [0,1]$, in order to make such a connection, we can create a relationship between these quantities by equating
\begin{equation*}
  \ln\left(\frac{\P(Y=y|x)}{1-\P(Y=y|x)}\right) = yp(w,x).
\end{equation*}
(The left-hand side of this equation is known as the \emph{logit} of $\P(Y=y|x)$.)  This implies, after some simple algebraic manipulations, that
\begin{equation}\label{eq.logit}
  \P(Y=y|x) = \frac{e^{yp(w,x)}}{1 + e^{yp(w,x)}} = \frac{1}{1 + e^{-yp(w,x)}},
\end{equation}
from which one can verify that the relationships in \eqref{eq.half} hold.  Following this idea, one finds that a reasonable surrogate for problem~\eqref{eq:expected_risk01} is
\begin{equation}\label{eq:expected_risk}
  \min_{w\in\Wcal}\ f(w),\ \ \text{where}\ \ f(w) := \int_{\Xcal \times \Ycal} \ell(p(w,x),y) dP(x,y) = \E[\ell(p(w,x),y)],
\end{equation}
where (by taking the negative logarithm of \eqref{eq.logit}) we use the \emph{logistic loss} function
\begin{equation*}
  \ell(p(w,x),y) = \log(1 + e^{-yp(w,x)}).
\end{equation*}
When $p(w,x)=w_0+w_1^T x$, this loss function is convex, and is, in fact, one of the most common loss functions used in learning to {\em train} linear predictors. Other convex loss functions that are commonly used are the {\em hinge loss} for classification, namely, $\ell(p(w,x),y) = \max\{0,1 -yp(w,x)\}$, and the \emph{least square loss} for regression, namely, $\ell(p(w,x),y)=\|p(w,x)-y\|^2$. In the case of linear predictors, these two loss functions typically give rise to convex quadratic optimization problems that may be very large-scale for which specific optimization algorithms have been designed.  (In this tutorial, we do not discuss such methods in detail since they are too specialized for our general setting.)

Problem~\eqref{eq:expected_risk} can be approached using stochastic optimization methods, such as stochastic approximation \cite{SpallBook,NemJudLanSha2009} and sample average approximation \cite{RuszShapiroBook,pasupathyinformstutorial}.  For guarantees in terms of solving~\eqref{eq:expected_risk}, the theory for such methods necessarily assumes that one can sample from $(\Xcal,\Ycal)$ indefinitely according to the distribution $P$.  However, in many settings in machine learning, this is not possible, so the last issue that we must confront is the fact that the objective of~\eqref{eq:expected_risk} and its derivatives cannot be computed since $P$ is unknown.  Instead, a surrogate problem can be considered.  For example, in \emph{supervised learning}, one assumes that there exists a set of input $\times$ output pairs (known as \emph{samples} or \emph{examples}) $\{(x_i,y_i)\}_{i=1}^n \subset (\Xcal,\Ycal)$ that is available \emph{a priori}.  With this set, one can approximately solve \eqref{eq:expected_risk} by solving a related \emph{deterministic} optimization problem defined over this set of samples, i.e.,
\begin{equation}\label{eq:empirical_risk}
  \min_{w\in\Wcal}\ \fhat(w),\ \ \text{where}\ \ \fhat(w) := \frac{1}{n} \sum_{i=1}^n \ell(p(w,x_i),y_i).
\end{equation}

Going forward, we will be interested in both problems~\eqref{eq:expected_risk} and \eqref{eq:empirical_risk}.  We consider the latter to be the problem that one can attempt to solve in practice, though we will also be interested in how solutions obtained from solving~\eqref{eq:empirical_risk} relate to optimal solutions of~\eqref{eq:expected_risk}.  We refer to $f$ as the \emph{expected risk} (\emph{loss}) function and refer to $\fhat$ as the \emph{empirical risk} (\emph{loss}) function.

\subsection{Learning bounds, overfitting, and regularization}\label{sec:learningbounds}

Let us now discuss relationships between problems~\eqref{eq:expected_risk} and \eqref{eq:empirical_risk} and their optimal solutions.  To start, suppose that these problems are convex, as is the case when $\ell$ is the logistic loss.  Let $w_*$ and $\hat w$, respectively, denote the expected and empirical risk minimizers.  Given that one aims for $\hat w$ minimizing $\fhat$ in hopes of approximating $w_*$ minimizing $f$, a variety of important questions arise about the potential differences between $\hat w$ and $w_*$, or between the optimal values of \eqref{eq:expected_risk} and \eqref{eq:empirical_risk}.  For example, one might be interested in a bound for
\begin{equation}\label{eq:error_sum}
  |\fhat(\what) - f(w_*)| \leq |\fhat(\what) - f(\what)| + |f(\what) - f(w_*)|.
\end{equation}
Bounding the first term on the right-hand side relates to asking: While $\what$ is designed to behave well on the sample set, can we be sure that it behaves well in expectation?  Bounding the second term relates to asking whether we can be sure that the expected loss corresponding to $\what$ is not far from the expected loss of the optimal vector with respect to $f$.


Different bounds can be derived for these quantities in various problem-specific cases.  Here, let us state a generic bound, which says that, with probability at least $1-\delta$, (see for example, \cite{vap95}) 
\begin{equation}\label{eq:estimerr}
  |\hat f(w)-f(w)| \leq \Ocal\left (\sqrt{\frac{C +\log(\frac{1}{\delta})}{n}}\right)\ \ \text{for}\ \ w \in {\cal W}, 
\end{equation}
where $C$ is a scalar representing the  \emph{complexity} of the class of the prediction functions, i.e., $p(w,\cdot)$ with $w \in {\cal W}$.  A similar bound can be derived on the second term in \eqref{eq:error_sum}.  From~\eqref{eq:estimerr}, a few intuitive notions are confirmed.  First, the discrepancy appears inversely proportional to $n$, meaning that more data leads the empirical risk to better approximate expected risk.  Second, the discrepancy depends on the complexity of the prediction functions.

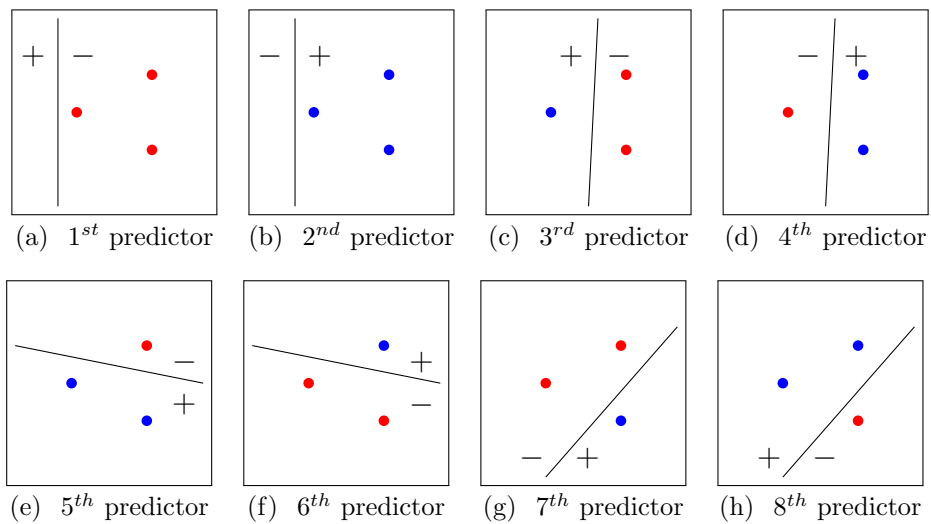
\begin{figure}[h]\hypertarget{mfl4}
\centering  

\subfigure[\ $1^{st}$ predictor]  
{  
\fbox{\begin{tikzpicture}[scale=.25]  
\path[use as bounding box] (0,0) rectangle (10,10);
\fill[red] (3,5) circle (8pt);
\fill[red] (7,3) circle (8pt);
\fill[red] (7,7) circle (8pt);
\draw (2,0)  -- (2,10); 
\node at (1.9,8) [left] {$+$};
\node at (2.1,8) [right] {$-$};

\end{tikzpicture}}
}  
\subfigure[\ $2^{nd}$ predictor]  
{  
\fbox{\begin{tikzpicture}[scale=.25]  
\path[use as bounding box] (0,0) rectangle (10,10);
\fill[blue] (3,5) circle (8pt);
\fill[blue] (7,3) circle (8pt);
\fill[blue] (7,7) circle (8pt);
\draw (2,0)  -- (2,10); 
\node at (1.9,8) [left] {$-$};
\node at (2.1,8) [right] {$+$};
\end{tikzpicture}}
}
\subfigure[\ $3^{rd}$ predictor]  
{  
\fbox{\begin{tikzpicture}[scale=.25]  
\path[use as bounding box] (0,0) rectangle (10,10);
\fill[blue] (3,5) circle (8pt);
\fill[red] (7,3) circle (8pt);
\fill[red] (7,7) circle (8pt);
\draw (5,0)  -- (5.5,10); 
\node at (5.3,8) [left] {$+$};
\node at (5.4,8) [right] {$-$};
\end{tikzpicture}}
}
\subfigure[\ $4^{th}$ predictor]  
{  
\fbox{\begin{tikzpicture}[scale=.25]  
\path[use as bounding box] (0,0) rectangle (10,10);
\fill[red] (3,5) circle (8pt);
\fill[blue] (7,3) circle (8pt);
\fill[blue] (7,7) circle (8pt);
\draw (5,0)  -- (5.5,10); 
\node at (5.3,8) [left] {$-$};
\node at (5.4,8) [right] {$+$};
\end{tikzpicture}}
}

\subfigure[\ $5^{th}$ predictor]  
{  
\fbox{\begin{tikzpicture}[scale=.25]  
\path[use as bounding box] (0,0) rectangle (10,10);
\fill[blue] (3,5) circle (8pt);
\fill[blue] (7,3) circle (8pt);
\fill[red] (7,7) circle (8pt);
\draw (0,7)  -- (10,5); 
\node at (9,5) [above] {$-$};
\node at (9,5) [below] {$+$};
\end{tikzpicture}}
}
\subfigure[\ $6^{th}$ predictor]  
{  
\fbox{\begin{tikzpicture}[scale=.25]  
\path[use as bounding box] (0,0) rectangle (10,10);
\fill[red] (3,5) circle (8pt);
\fill[red] (7,3) circle (8pt);
\fill[blue] (7,7) circle (8pt);
\draw (0,7)  -- (10,5); 
\node at (9,5) [above] {$+$};
\node at (9,5) [below] {$-$};
\end{tikzpicture}}
}
\subfigure[\ $7^{th}$ predictor]  
{  
\fbox{\begin{tikzpicture}[scale=.25]  
\path[use as bounding box] (0,0) rectangle (10,10);
\fill[red] (3,5) circle (8pt);
\fill[blue] (7,3) circle (8pt);
\fill[red] (7,7) circle (8pt);
\draw (3,0)  -- (10,8); 
\node at (3.5,1) [left] {$-$};
\node at (4,1) [right] {$+$};
\end{tikzpicture}}
}
\subfigure[\ $8^{th}$ predictor]  
{  
\fbox{\begin{tikzpicture}[scale=.25]  
\path[use as bounding box] (0,0) rectangle (10,10);
\fill[blue] (3,5) circle (8pt);
\fill[red] (7,3) circle (8pt);
\fill[blue] (7,7) circle (8pt);
\draw (3,0)  -- (10,8); 
\node at (3.5,1) [left] {$+$};
\node at (4,1) [right] {$-$};
\end{tikzpicture}}
}
\caption{Correct classifications of all 8 possible labelings of 3 points by linear predictors in $\R^2$}
\label{fig:VCdimR2}
\end{figure}

It is beyond the scope of this tutorial to show how one might derive~$C$ in general.  Instead, for the sake of intuition, let us briefly introduce one such measure for binary classification: the Vapnik-Chervonenkis (VC) dimension.  Briefly, the VC-dimension $C$ of a class of predictors $p(w,\cdot)$ with $w\in {\cal W}$ is the largest number for which there exists a set of points $\{x_1, x_2, \ldots, x_C\}$ such that for any label set $\{y_1, y_2, \ldots, y_C\}$, there exists a predictor $p(w,x)$ that predicts all labels without error, i.e., $y_ip(w,x_i)>0$ for all $i \in \{1,\dots,C\}$.  For example, for the class of linear predictors of the form $p(w,x) = w_0+ w_1^Tx$ with $(w_0,w_1) \in \R^{m+1}$, the VC-dimension is 
$m+1$.  To see this, in Figure \ref{fig:VCdimR2} we illustrate a set of 3 points in $\R^2$ and a linear predictor for each labeling of these points, where in each case one finds a predictor that predicts all labels without error.  On the other hand, it is easy to show that for some set of $4$ distinct points in~$\R^2$, there exists at least one labeling of the points which cannot be perfectly classified by a linear function.  Hence, the VC-dimension of linear predictors in $\R^2$ is $3$.

For linear predictors, the VC-dimension is equal to the number of parameters that define the prediction function.  However, this is not true for all predictor classes.  In such cases, instead of using a precise measure of complexity in \eqref{eq:estimerr}, a bound can be used.  For example, if the input space $\Xcal$ is bounded by an $\ell_2$-norm ball of radius $R_x$ and the class of predictors $p(w,x)$ is such that $w$ is constrained in a ball of radius $R_w$, then, with smoothness of a loss function $\ell(\cdot, \cdot)$,  a bound on $C$ can be derived in terms of $\Ocal(R_x^2R_w^2)$. For more information on complexity measures of classes of predictors, we refer the reader to \cite{BartlettMandel,GneccoSanguineti2008,vap95}.

For our purposes going forward, it is important to observe that as $C$ gets larger, a larger sample set is required to keep the right-hand side in \eqref{eq:estimerr} small.  This means that in order for the optimal value of problem~\eqref{eq:empirical_risk} to represent the actual expected risk corresponding to $\hat w$, there needs to be a balance between the complexity of the predictor class and the number of sample points. If optimization is performed over a complex class of predictors using a sample set that is not sufficiently large, then the optimal solution $\hat w$ may achieve small $\hat f(\hat w)$, but have a large expected loss $f(\hat w)$. Such a solution will have poor predictive properties because it \emph{overfits} the training data.  On the other hand, to learn ``effectively'', one needs to balance the complexity of the predictors and the size of the data set.

One way to control the complexity of a class is simply to control the $\ell_2$-norm of $w$, since smaller $\|w\|_2$ results in smaller $R_w$ and a smaller bound on an appropriate complexity measure $C$ \cite{BartlettMandel}.  Alternatively, particularly for linear predictors, one can control complexity by bounding the number of nonzeros in $w$, thereby constraining the class to sparse predictors.  This idea is of particular interest when each data vector $x$ has a lot of features, yet the prediction function $w_0 + w_1^Tx$ is believed to depend only on a small (unknown) subset of these features.  Once the subset of features is selected, the VC-dimension of the class reduces to the number of nonzeros in $w$.  Hence, if a sparse predictor achieves small empirical error, then it is likely to achieve small expected error.

Instead of explicitly constraining the number of nonzeros in $w$, which would make the optimization problem harder, a regularization term such as $\lambda \|w\|_1$ can be added to the objective function.  The addition of this term encourages {\em feature selection}.  It does not always guarantee that a sparse solution will be reached, but it has been shown to work well in practice.  
Generically, to attempt to restrict the complexity of the prediction function class in some manner, one often considers a regularized optimization problem of the form 
\begin{equation}
\label{eq:P}
  \min_{w \in {\bf R}^d}\ F(w),\ \ \text{where}\ \ F(w) = \frac{1}{n} \sum_{i=1}^n \ell(p(w,x_i), y_i) + \lambda r(w),
\end{equation}
$\lambda \geq 0$ is a weighting parameter, and $r$ is either $\|w\|_1$ or some other convex (potentially nonsmooth) regularization function.  How the parameter $\lambda$ should be chosen is a subject of work in {\em structural risk minimization}.  However, the details of this are beyond the scope of this tutorial.  For our purposes, suffice it to say that for any given value of $\lambda$, one has an optimization problem to solve (at least approximately), so now let us turn to optimization algorithms that may serve well in the context of problem~\eqref{eq:P}.

\section{Methods for Solving the Logistic Regression Problem}\label{sec:Logreg}

The methods that we discuss in this section for solving problem~\eqref{eq:P} could be employed when $\ell$ and $r$ are any convex functions with respect to~$w$. 
There is a large variety of machine learning models that fall in this category, including support vector machines, Lasso, sparse inverse covariance selection and others.
For further details on these models  see \cite{Wrightsurvey} and references therein. Here, in order to be more concrete at various times, we will  refer specifically to the case of regularized logistic regression for binary classification.  To simplify our notation in this case, let us assume without loss of generality that $p(w,x) = w^Tx$.  (That is, we omit the bias term $w_0$, which can be done by augmenting the input vector by an extra feature which is always equal to $1$.)  Denoting the dimension of both $w$ and $x$ as $d$, this leads to the specific convex problem
\begin{equation}\label{eq:minempreglogloss}
  \min_{w \in {\bf R}^d} F(w),\ \ \text{where}\ \ F(w) = \frac{1}{n} \sum_{i=1}^n \log \left(1+e^{-y_i(w^Tx_i)}\right) + \lambda r(w).
\end{equation}
It is worthwhile to note that the regularization term is necessary for this problem.  To see why this is the case, consider a parameter vector $w$ such that $y_i(w^Tx_i) > 0$ for all $i \in \{1,\dots,n\}$.  Then, consider the unbounded ray $\{\theta w : \theta > 0\}$.  It is easy to see that, in this case, $\frac{1}{n}\textstyle{\sum}_{i=1}^n \log(1+e^{-y_i(\theta w^Tx_i)}) \to 0$ as $\theta\to \infty$, meaning that the minimum of this function cannot be achieved.  On the other hand, by adding a (coercive) regularization function $r$, it is guaranteed that problem \eqref{eq:minempreglogloss} will have an optimal solution.

For the regularization function $r$, we will refer to the common choices of $r(w) = \|w\|_2^2$ and $r(w) = \|w\|_1$.  For simplicity, we will refer mostly to the former choice, which makes the objective of \eqref{eq:minempreglogloss} a continuously differentiable function.  By contrast, $r(w) = \|w\|_1$ results in a nonsmooth problem, for which minimization requires more sophisticated algorithms.

\subsection{First-order methods}

We begin by discussing first-order methods for solving~\eqref{eq:minempreglogloss}.  Here, ``first-order'' refers to the fact that these techniques only require first-order derivatives of the terms in $F$.

\subsubsection{Gradient descent}\label{sec.gradient_descent}

Conceptually, the most straightforward method for minimizing a smooth convex objective is \emph{gradient descent}; e.g., see~\cite{NocedalWright06}.  In this approach, one starts with an initial solution estimate $w_0$
and iteratively updates the estimate via the formula
\begin{equation}\label{eq.gradient_descent}
  w_{k+1} \gets w_k - \alpha_k \nabla F(w_k),
\end{equation}
where $\alpha_k > 0$ is a stepsize parameter.  The performance of the algorithm is inherently tied to the choice of stepsize sequence $\{\alpha_k\}$.  In the optimization research community, it is well known that employing a \emph{line search} in each iteration to determine $\{\alpha_k\}$ can lead to a well-performing algorithm for a variety of types of problems.  However, for ML applications, such operations are expensive due to the fact that each computation of $F$ requires a pass over the entire dataset, which can be prohibitively expensive if $n$ is large.

Fortunately, theoretical convergence guarantees for the algorithm can still be proved when each~$\alpha_k$ is set to a positive constant $\alpha$ for all $k$, as long as this fixed value is sufficiently small.  (When the stepsize is fixed, it is known in the ML community as the \emph{learning rate} for the algorithm.  Some also use this term to refer to each $\alpha_k$ or the entire sequence $\{\alpha_k\}$, even when it is not constant.)  The convergence rate depends on whether $F$ is strongly convex or merely convex.  If $F$ is $\mu$-strongly convex, the gradient function $\nabla F$ is Lipschitz continuous with Lipschitz constant $L \geq \mu$, and $\alpha \in (0,1/L)$, then it can be shown that the number of iterations required until $F(w_k) - F(w_*) \leq \epsilon$ is at most $\Ocal(\kappa \log(1/\epsilon))$, where $w_* := \arg\min_w F(w)$ and $\kappa := L/\mu$.  This is a \emph{linear} rate of convergence.  (If $\lambda > 0$, then these conditions hold for problem~\eqref{eq:minempreglogloss}; in particular, $\mu \geq \lambda$ and $L \leq \max_{i\in\{1,\dots,n\}} \|x_i\|_2^2$.)  On the other hand, if $F$ is merely convex (as would be the case in \eqref{eq:minempreglogloss} when $\lambda=0$), then such an $\epsilon$-optimal solution is obtained within at most $\Ocal(1/\epsilon)$ iterations, which is a \emph{sublinear} rate.  One can actually improve these rates by employing an \emph{acceleration} technique by Nesterov~\cite{Nesterov}.  Using such an approach allows one to attain $\epsilon$-optimality within $\Ocal\left(\sqrt{\kappa} \log\left(1/\epsilon\right)\right)$ iterations when~$F$ is strongly convex and within $\Ocal(1/\sqrt{\epsilon})$ iterations when it is convex.

Extensions of gradient descent and accelerated gradient descent for solving $\ell_1$-norm regularized logistic regression problems (i.e., \eqref{eq:minempreglogloss} when $r(w) = \|x\|_1$) are known as ISTA and FISTA, respectively, \cite{Beck-Teboulle-2009}.  Observe that, in this setting, the objective function may not be strongly convex even if $\lambda > 0$.  ISTA and FISTA have the same sublinear convergence rates as their smooth counterparts when the objective function is convex \cite{Beck-Teboulle-2009}.


The important aspect of gradient descent in the context of ML is to recognize the computational cost of computing the gradient of $F$ in each iteration.  In particular, in the context of ML, the cost of a single gradient computation is typically $\Ocal(nd)$; this can be seen, e.g., in the case of problem~\eqref{eq:minempreglogloss} with $r(w) = \|w\|_2^2$, where the gradient of $F$ for a given $w$ is
\begin{equation}\label{eq:loglossgrad}
  \nabla F(w) = -\frac{1}{n} \sum_{i=1}^n \left(\frac{1}{1 + e^{y_i(w^Tx_i)}}\right)y_ix_i + 2\lambda w.
\end{equation}
(Here, one funds a sum of $n$ terms, where for each term one needs to compute the inner product $w^Tx_i$ of $d$-dimensional vectors, hence the $\Ocal(nd)$ cost.)  The dependence of this computation on $n$ (which can be of order $10^6\sim 10^9$ in various ML applications) is computationally prohibitive, and can be viewed as potentially wasteful when many of the elements of the sample set are the same or at least very similar.  In the next subsection, we discuss a \emph{stochastic} optimization method whose per-iteration computational cost is drastically smaller, and yet can still offer convergence guarantees.


\subsubsection{Stochastic gradient method}

The stochastic gradient method, well known in the OR community due to its use for minimizing stochastic objective functions, is the hallmark optimization algorithm in the ML community.  Originally proposed by Robbins and Monro \cite{Robbins:1951ko} in the context of solving stochastic systems of equations, the method is notable in that it can be employed to minimize a stochastic objective with nice convergence guarantees while the per-iteration cost is only $\Ocal(d)$ as opposed to $\Ocal(nd)$ (as in gradient descent). 

In each iteration, the stochastic gradient method computes an unbiased estimator $G_k$ of the true gradient $\nabla F(w_k)$.  This estimator can be computed at very low cost; e.g., for~\eqref{eq:minempreglogloss}, a stochastic gradient can be computed as
\begin{equation}\label{eq:sgd}
  \nabla_{S_k}F(w) = -\frac{1}{|S_k|} \sum_{i\in S_k} \left(\frac{1}{ 1+e^{ y_i (w^T x_i)} }\right)y_ix_i + 2\lambda w,
\end{equation}
where $S_k$, known as the \emph{mini-batch}, has elements chosen uniformly at random from $\{1, \ldots, n\}$.  The step is then taken similar to gradient descent:
\begin{equation}\label{eq.sgd_step}
  w_{k+1} \gets w_k - \alpha_k \nabla_{S_k}F(w_k).
\end{equation}
Absolutely critical to the algorithm is the choice of the stepsize sequence $\{\alpha_k\}$.  Unlike gradient descent, a fixed stepsize (i.e., learning rate) does not guarantee convergence of the algorithm to the minimizer of a strongly convex $F$, but rather only guarantees convergence to a neighborhood of the minimizer.  However, Robbins and Monro showed that with
\begin{equation*}
  \sum_{k=1}^\infty \alpha_k = \infty\ \ \text{and}\ \ \sum_{k=1}^\infty \alpha_k^2<\infty,
\end{equation*}
one can guarantee a sublinear rate of convergence (almost surely) to the minimizer \cite{BottCurtNoce16}.  Note that the stochastic gradient method is not, strictly speaking, a descent method since $-\nabla_{S_k}F(w_k)$ is not guaranteed to be a descent direction for $F$ from $w_k$ and the objective function does not decrease monotonically over the optimization process.  However, we will refer to it as SGD, for \emph{stochastic gradient descent}, as is often done in the literature. 

The convergence rate of SGD is slower than that of gradient descent.  In particular, when $F$ is strongly convex, it only guarantees that an expected $\epsilon$-accurate solution (i.e., one with $\E[F(w_k)] - F(w^*)\leq \epsilon$) is obtained once $k \geq \Ocal(1/\epsilon)$, and when $F$ is merely convex, it only guarantees that such a solution is found once $k \geq \Ocal(1/\epsilon^2)$ \cite{BottCurtNoce16}.  On the other hand, as previously mentioned, if the size of $S_k$ is bounded by a constant (independent of $n$ or $k$), then the per-iteration cost of SGD is $\Ocal(n)$ times smaller than that of gradient descent.

This trade-off between convergence rate and per-iteration cost can be analyzed in the context of ML, with strong results in favor of SGD; e.g., see~\cite{BottouBousquet}.  To outline the ideas behind this analysis, let us ignore regularization terms and recall that our ultimate goal is to solve~\eqref{eq:expected_risk} with some accuracy $\epsilon$.  This means that we want to obtain $w_\epsilon$ such that $f(w_\epsilon)-f(w_*)\leq \epsilon$, where $w_*$ is the optimal solution to  \eqref{eq:expected_risk}.  Instead, however, we solve problem  \eqref{eq:empirical_risk} to some accuracy $\hat\epsilon$, obtaining $\what_{\hat\epsilon}$ such that $\E[\hat f(\what_{\hat\epsilon})-\hat f(\hat w)] \leq \hat\epsilon$, where $\what$ is the optimal solution to \eqref{eq:empirical_risk}.  If we use $\what_{\hat\epsilon}$ as our approximate solution to \eqref{eq:expected_risk}, then, from~\eqref{eq:error_sum} and \eqref{eq:estimerr} and since $\E[\fhat(\what) - \fhat(w_*)] \leq 0$, we have with probability at least $1-\delta$ that
\begin{align}
  \E[f(\what_{\hat\epsilon}) - f(w_*)]
       =&\ \E[f(\what_{\hat\epsilon}) - \fhat(\what_{\hat\epsilon})] + \E[\fhat(\what_{\hat\epsilon}) - \fhat(\what)] \nonumber \\
        &\ + \E[\fhat(\what) - \fhat(w_*)] + \E[\fhat(w_*) - f(w_*)] \nonumber \\
    \leq&\ \Ocal\left(\sqrt{\frac{C + \log(\tfrac{1}{\delta})}{n}}\right) + \hat\epsilon. \label{eq:BottBous}
\end{align}
Thus, to achieve expected $\epsilon$-optimality with respect to~\eqref{eq:expected_risk} while balancing the contributions of the terms on the right-hand side of~\eqref{eq:BottBous}, we should aim to have, say,
\begin{equation}\label{eq.bounds_needed}
  \Ocal\left(\sqrt{\frac{C +\log(\frac{1}{\delta})}{n}}\right)\leq \frac{\epsilon}{2}\ \ \text{and}\ \ \hat\epsilon \leq \frac {\epsilon}{2}.
\end{equation}

We can now compare algorithms by quantifying the computational costs they require to satisfy these bounds.  For example, suppose that we apply some algorithm to solve problem~\eqref{eq:empirical_risk}, where, for a given $n$, the cost of obtaining $\what_{\hat\epsilon}$ satisfying the latter bound in \eqref{eq.bounds_needed} is $c(n,\epsilon)$, which increases with both $n$ and $1/\epsilon$.  For a fixed family of functions $p(w,\cdot)$ with $w \in \Wcal$ and complexity $C$, obtaining the former bound in \eqref{eq.bounds_needed} requires $n \geq \Ocal(\epsilon^2)$ (ignoring log factors).  Considering now the optimization algorithm and its cost $c(n,\epsilon)$, it is clear that any algorithm that computes $\fhat$, its gradient, or its Hessian at any point has a cost of at least $\Ocal(n) = \Ocal(1/\epsilon^2)$ \emph{to perform a single iteration}, regardless of the rate at which it converges to a minimizer of $\fhat$.  This is the case, e.g., for the gradient descent method.  SGD, on the other hand, has a per-iteration cost that is independent of $n$ and can be shown to \emph{converge to an $\tfrac{\epsilon}{2}$-optimal solution} within at most $\Ocal(1/\epsilon^2)$ iterations (when the objective function is convex and one can sample indefinitely from the input $\times$ output space according to the distribution $P$).  From this discussion, we can conclude that, at least in theory, SGD is a superior algorithm for large-scale (i.e., large $n$) ML applications.


In practice, however, standard SGD is not necessarily the most effective approach to solve optimization problems in ML.  Indeed, there is a great deal of active research in the ML and optimization communities on developing improvements and/or alternatives to SGD.  In the subsequent two sections, we discuss two categories of such methods: \emph{variance reducing} and \emph{second-order methods}.  However, there are a variety of approaches even beyond these two categories.  For example, one extension of SGD that has often been found to perform better in practice is SGD \emph{with momentum}; see Algorithm~\ref{alg:msgd}.  This algorithm is a stochastic variant of the classical momentum method by Polyak~\cite{Pol64}, which enjoys an improved convergence rate compared to standard gradient descent.  SGD with momentum has been shown to be quite successful in machine learning, especially deep learning, our topic in~\S\ref{sec.dnn}; e.g., see~\cite{Sutskeveretal2013}. It is shown in \cite{Sutskeveretal2013} that Nesterov's accelerated gradient method~\cite{Nesterov} can be cast as a classical  momentum approach.  That said, to the best of our knowledge, this stochastic version of the momentum method does not have convergence guarantees.  Some results analyzing stochastic variants of acceleration methods can be found in~\cite{Lan2012}. 

\begin{algorithm}
  \caption{SGD with Momentum}
  \label{alg:msgd}
  \begin{algorithmic}
    \State {\bfseries Parameters:} learning rate $\alpha > 0$; momentum weight $\eta \in (0,1)$; mini-batch size $s \in \mathbb{N}$
    \State {\bfseries Initialize:} $w_0 \in \R^n$; $v_0 = 0 \in \R^n$
    \State {\bfseries Iterate:}
    \For{$k=1,2,\dots$}
      \State Generate $S_k$ with $|S_k|=s$ uniformly from $\{1,\dots,n\}$ 
      \State Compute $\nabla_{S_k}F(w_k)$ according to \eqref{eq:sgd}
      \State Set $v_k \gets \eta v_{k-1} +(1-\eta) \nabla_{S_k}F(w_k)$
      \State Set $w_{k+1} \gets w_k - \alpha v_k$
    \EndFor
  \end{algorithmic}
\end{algorithm}

As a final remark on SGD, we note that aside from its slow convergence rate (in theory and practice), SGD is very sensitive to parameter choices, such as the mini-batch size $|S_k|$ and learning rate.  The best choice of these parameters heavily depends on the dataset, and the wrong choice can severely slow progress or cause the algorithm to stall completely.

\subsubsection{Variance reducing methods}

The arguments for SGD that we raised in the previous section rely on the assumptions that the true aim is to solve problem~\eqref{eq:expected_risk} and that one can sample indefinitely from the space of inputs and outputs.  However, in ML applications, the sample set $\{(x_i,y_i)\}_{i=1}^n$ is often given and fixed, and if $n$ is sufficiently large, then there is good reason to view a discrete uniform distribution over the sample set as a good approximation to the distribution $P$.  Thus, one can argue that the conclusions of the previous section should only be used as a theoretical guideline, reminding us that while \eqref{eq:expected_risk} might be the true problem of interest, one can reasonably be interested in the most efficient methods for solving \eqref{eq:empirical_risk}, or, very often, the regularized problem~\eqref{eq:P}.

Considering problem~\eqref{eq:P}, one finds that SGD can be improved upon by exploiting the structure of the objective $F$ as a finite
sum of $n$ functions plus a simple convex term.  Several methods have been developed along these lines, such as SAG \cite{schmidt2013minimizing}, SAGA \cite{NIPS2014_5258}, SDCA \cite{shalev2013accelerated}, and SVRG \cite{johnson2013accelerating}.  SAG and SAGA, for example, rely on averaging the past $n$ stochastic gradients in a particular manner in an attempt to accumulate more accurate gradient estimates as the optimization algorithm proceeds. As
a result, they enjoy the same convergence rate as full gradient methods (with better constant factors).  However, these methods require the storage of $n$ past stochastic gradients so that components can be individually updated as the method progresses.  In contrast, SVRG does not require such storage, though it does require computing the full gradient every $\Ocal(n)$ iterations.  For reference, we state SVRG as Algorithm~\ref{alg:svrg}. The algorithm performs one full gradient computation at each outer iteration, then takes $l$ steps along random directions which are stochastic corrections of this full gradient.  The inner loop size $l$ must satisfy certain conditions to ensure convergence~\cite{johnson2013accelerating}. 

\begin{algorithm}
  \caption{SVRG}
  \label{alg:svrg}
  \begin{algorithmic}
    \State {\bfseries Parameters:} learning rate $\alpha > 0$; mini-batch size $s \in \mathbb{N}$; inner loop size $l \in \mathbb{N}$
    \State {\bfseries Initialize:} $\tilde{w}_0 \in \R^n$
    \State {\bfseries Iterate:}
    \For{$k=1,2,\dots$}
      \State Set $w_0 \gets \tilde{w}_{k-1}$
      \State Set $v_0 \gets \nabla F(w_0)$
      \State Set $w_1 \gets w_0 - \alpha v_0$
      \State {\bfseries Iterate:}
      \For{$t=1,\dots,l-1$}
        \State Generate $S_t$ with $|S_t| = s$ uniformly from $\{1,\dots,n\}$
        \State Compute $\nabla_{S_t}F(w_t)$ and $\nabla_{S_t}F(w_0)$ according to \eqref{eq:sgd}
        \State Set $v_t \gets \nabla_{S_t}F(w_t) - \nabla_{S_t} F(w_0) + v_0$
        \State Set $w_{t+1} \gets w_{t} - \alpha v_t$
      \EndFor
      \State Set $\tilde{w}_k = w_t$ with $t$ chosen uniformly from $\{0,\dots,l\}$
    \EndFor
  \end{algorithmic}
\end{algorithm}

The name SVRG, which stands for \emph{stochastic variance-reduced gradient}, comes from the fact that the algorithm can be viewed as a variance-reduced variant of SGD (specifically for finite-sum minimization).  To see this, first observe that the random directions taken by the algorithm are in fact unbiased gradient estimates:
\begin{equation}\label{eq:SVRGisSGD}
  \E[v_t] = \nabla F(w_t) - \nabla F(w_0) + v_0 = \nabla F(w_t).
\end{equation}
Second, notice that computation of the full gradient at each outer iteration helps reduce the variance of the stochastic gradient estimates used in SGD; indeed, if $w_t$ is ``close'' to~$w_0$, then one should expect $v_t$ to be ``closer'' to $\nabla F(w_t)$ than is $\nabla_{S_t} F(w_t)$ alone.  In practice, SVRG is somewhat more robust than SGD is to the choice of learning rate $\alpha$, though it still can be quite sensitive to the choice of $l$, the inner loop size. 

A new method that combines some ideas from SVRG and SAGA, called SARAH \cite{Sarah2017}, only differs from SVRG in terms of the inner loop step,  which in SARAH is given by 
\begin{equation}\label{eq:vt}
   v_{t} \gets \nabla_{S_t} F(w_t) - \nabla_{S_t} F(w_{t-1}) + v_{t-1}.
\end{equation}
This change causes $\E[v_t] \neq \nabla F(w_{t})$ so that the steps in SARAH are not based on unbiased gradient estimates.  However, it attains improved convergence properties relative to SVRG.

In Tables~\ref{table:summary} and~\ref{table:summary_convex}, we summarize the complexity properties of several popular first-order methods when applied to minimize strongly convex (Table~\ref{table:summary}) and convex (Table~\ref{table:summary_convex}) problems.  Here, complexity refers to an upper bound on the number of iterations that may be required to attain an $\epsilon$-accurate solution.  In the former case, recall $\kappa := L/\mu$ defined in \S\ref{sec.gradient_descent}, often referred to as the condition number of a strongly convex problem.

\begin{table}[ht]
  \small
  \caption{Complexity of first-order methods when minimizing strongly convex functions}
  \label{table:summary}
  \centering
  \begin{tabular}{cc}
    Method & Complexity   \\
    \hline
    GD & $\Ocal\left(n\kappa \log\left({1}/{\epsilon}\right)\right)$  \\
    \hline
    SGD & $\Ocal\left({1}/{\epsilon}\right)$   \\
    \hline
    SVRG & $\Ocal\left((n+\kappa) \log\left({1}/{\epsilon}\right)\right)$    \\
    \hline
    SAG/SAGA & $\Ocal\left((n + \kappa) \log\left({1}/{\epsilon}\right)\right)$   \\
\hline
    SARAH & {$\Ocal\left((n + \kappa) \log\left({1}/{\epsilon}\right)\right)$}  \\
    \hline
  \end{tabular}
  \vskip0.3cm

  \caption{Complexity of first-order methods when minimizing (not strongly) convex functions}
  \label{table:summary_convex}
  \begin{tabular}{cc}
    Method & Complexity  \\
    \hline
    GD & $\Ocal\left(n/\epsilon \right)$  \\
    \hline
    SGD & $\Ocal\left({1}/{\epsilon^2}\right)$  \\
    \hline
    SVRG & $\Ocal\left(n + (\sqrt{n}/\epsilon) \right)$   \\
    \hline
    SAGA & $\Ocal\left(n + (n/\epsilon) \right)$  \\
    \hline
    SARAH & {$\Ocal\left((n + (1/\epsilon)) \log(1/\epsilon) \right)$} \\
    \hline
  \end{tabular}
\end{table}

Yet another branch of variance reducing methods are those that employ the standard SGD step formula~\eqref{eq.sgd_step}, but attempt to achieve a faster convergence rate by increasing the mini-batch size during the optimization process.  If done carefully (and one can sample indefinitely from the input $\times$ output space), then such an approach can achieve a linear rate of convergence for minimizing the expected risk~\eqref{eq:expected_risk}; e.g., see \cite{BottCurtNoce16}.  Similar ideas have also been explored for the case of minimizing finite sums; e.g., see \cite{doi:10.1137/110830629}.

\subsection{Second-order and quasi-Newton methods}\label{sec:bfgs}

Motivated by decades of work in the deterministic optimization research community, one of the most active research areas in optimization for ML relates to how one might use second-order derivative (i.e., curvature) information to speed up training.  As in the deterministic setting, the idea is to compute the $k$th step by approximately minimizing a quadratic model of the objective function $F$ about $w_k$ of the form
\begin{equation}\label{eq:quadmod}
  m_k(s) = F(w_k) + \nabla F(w_k)^T s + \tfrac{1}{2} s^TB_ks,
\end{equation}
where $B_k$ is positive definite, i.e., $B_k \succ 0$.  A variety of algorithms of this type exist, each distinguished by how the curvature matrix~$B_k$ is obtained and how an (approximate or exact) minimizer~$s_k$ is computed.  The prototypical example is Newton's method where $B_k = \nabla^2 F(w_k)$, assuming this matrix is positive (semi)definite. For example, for the case of the regularized logistic regression problem~\eqref{eq:minempreglogloss}, one finds that this Hessian matrix, like the gradient in~\eqref{eq:loglossgrad}, comes from the sum of similar terms defined over the dataset:
\begin{equation}\label{eq:logregHess}
  \nabla^2 F(w) = \frac{1}{n}\sum_{i=1}^n \left(\frac{e^{y_i(w^Tx_i)}}{(1+e^{y_i(w^T x_i)})^2} x_ix_i^T\right) + 2\lambda I_{d \times d}.
\end{equation}

Unfortunately, the computation and storage of a Hessian matrix becomes prohibitively expensive in ML applications when $n$ and/or $d$ is large.  As an alternative, one may consider using stochastic Hessian information by replacing the average in \eqref{eq:logregHess} with an average over a small subset $S_k \subseteq \{1,\dots,n\}$, as is done for computing gradient estimates.  In the case of \eqref{eq:logregHess}, this might appear to be a good idea since the stochastic Hessian is a sum of $|S_k|$ rank-one matrices and a scaled identity matrix, meaning that solving a system of equations with such a matrix might not be too expensive.  However, such a low-rank approximation might not adequately capture complete curvature information, making the added costs of the algorithm (as compared to those of SGD or one of its variants) not worthwhile. Recently, several variants of {\em sub-sampled} Newton methods have been proposed, where the sample size $|S_k|$ is increased to improve the accuracy of the Hessian estimates as the algorithms progresses.  We give a brief overview of these ideas in~\S \ref{sec:SH}.

Another class of algorithms based on models of the form~\eqref{eq:quadmod} are \emph{quasi-Newton} methods.  Interestingly, these approaches compute \emph{no explicit} second-order derivatives; instead, they construct Hessian approximation matrices entirely from first-order derivatives by applying low-rank updates in each iteration.  For example, let us briefly describe the most popular quasi-Newton algorithm, known as the Broyden-Fletcher-Goldfarb-Shanno (BFGS) method.  In this approach, one first recognizes that the minimizer of \eqref{eq:quadmod} is $-B_k^{-1}\nabla F(w_k)$, revealing that it is actually convenient to compute \emph{inverse} Hessian approximations.  With $B_k^{-1}$ in hand along with the step $s_k=w_{k+1}-w_k$ and displacement $y_k=\nabla F(w_{k+1})-\nabla F(w_k)$, one chooses $B_{k+1}^{-1}$ to minimize $\|B^{-1} - B_k^{-1}\|$ subject to satisfying the \emph{secant equation} $s_k = B^{-1}y_k$.  Using a carefully selected norm, the solution of this problem can be written explicitly as
\begin{equation}\label{eq:bfgsinv}
  B^{-1}_{k+1} = \left(I-\frac{s_ky_k^T}{s_k^Ty_k}\right) B^{-1}_k \left(I-\frac{y_ks_k^T}{y_k^T s_k}\right) + \frac{s_ks_k^T}{y_k^Ts_k},
\end{equation}
where the difference between $B_k^{-1}$ and $B_{k+1}^{-1}$ can be shown to be only a rank-two matrix.

For reference, a complete classical BFGS algorithm is stated as Algorithm~\ref{alg:bfgs}. 

\begin{algorithm}[ht]
  \caption{BFGS method}
  \label{alg:bfgs}
  \begin{algorithmic}
    \State {\bfseries Initialize:} ${w}_0 \in \R^n$; $B_0^{-1} \in \R^{n \times n}$ with $B_0^{-1} \succ 0$.
    \State {\bfseries Iterate:}
    \For{$k = 0,1,\dots$}
      \State Set $\tilde{s}_k \gets B_k^{-1}\nabla F(w_k)$
      \State Compute $\alpha_k$ satisfying the Wolfe line search conditions \cite{NocedalWright06} for $F$ from $w_k$ along $\tilde{s}_k$
      \State Set $s_k \gets \alpha_k\tilde{s}_k$
      \State Set $y_k \gets \nabla F(w_k+s_k) - \nabla F(w_k)$
      \State Set $B_{k+1}^{-1}$ by \eqref{eq:bfgsinv}
      \State Set $w_{k+1} \gets w_k + s_k$
    \EndFor
  \end{algorithmic}
\end{algorithm}

It has been shown (e.g., see \cite{NocedalWright06}), under certain conditions including strong convexity of the objective function $F$, that the BFGS method yields the Dennis-Mor\'e condition, namely, $\lim_{k\to \infty} \|(B_k-\nabla ^2 F(w_k))s_k\| /\|s_k\|=0$, which in turn can be used to show that the BFGS method converges locally superlinearly.  Critical in this analysis is the condition that $s_k^Ty_k > 0$ for all $k$, which guarantees that $B_{k+1}$ in \eqref{eq:bfgsinv} is positive definite as long as $B_k \succ 0$.  When the objective $F$ is strongly convex, this holds automatically, but, for merely convex or potentially nonconvex $F$, this is guaranteed by the Wolfe line search.

In the context of large-scale ML, the classical BFGS method is intractable for a variety of reasons.  Of course, there are the issues of the computational costs of computing exact gradients and performing a line search in each iteration.  However, in addition, there is the fact that the Hessian approximations will almost certainly be dense, which might lead to storage issues in addition to the expense of computing matrix-vector products with large dense matrices.  Fortunately, however, the optimization literature has already addressed these latter issues with the idea of using \emph{limited memory} quasi-Newton approaches, such as limited memory BFGS, also known as L-BFGS~\cite{ByrdNoceSchn1994}.


For our purposes, let us simply give an overview of several variants of the (L-)BFGS method that have been recently developed to address various challenges of ML applications.  One common feature of all of these approaches is that the line search component is completely removed in favor of a predetermined stepsize sequence such as is used for SGD.  Another common feature is that the sequence $\{y_k\}$ no longer represents (exact) gradient displacements, since exact gradients are too expensive to compute.  Instead, the techniques mentioned in the bullets below use different ideas for this sequence of vectors.

\begin{itemize}
\item
  If one defines $y_k$ as the difference of two stochastic gradient estimates formed using two different mini-batches, i.e., $y_k=\nabla _{S_{k+1}}F(w_{k+1})-\nabla _{S_{k}}F(w_{k})$, then one typically fails to have $s_k^Ty_k > 0$ and, even if this bound does hold, the Hessian approximation sequence can be highly volatile.  An alternative idea that overcomes these issues is to definte $y_k = \nabla _{S_{k}}F(w_{k+1})-\nabla _{S_{k}}F(w_{k})$ so that the displacement is computed using the \emph{same} mini-batch at each point.  This is what is proposed in the methods known as oBFGS (oLBFGS)~\cite{Schraudolphetal} and    SGD-QN \cite{Bordesetal}.  This idea requires two computations of a stochastic gradient per iteration, one for each of the mini-batches $S_{k-1}$ (to compute the previous displacement) and~$S_k$ (to compute the next step). The resulting method has been shown to perform well for convex problems, though convergence eventually slows down when noise in the gradient estimates starts to dominate the improvements made by the steps.
  \item 
In \cite{BeraNoceTaka2016}, it is proposed to use   large \emph{overlapping} (but still distinct) mini-batches on consecutive iterations within an L-BFGS framework. 
In other words, $y_k=\nabla _{S_{k+1}}F(w_{k+1})-\nabla _{S_{k}}F(w_{k})$, with $S_{k+1}$ and $S_k$ not independent, but containing a relatively large overlap. 
Positive results on optimizing the training error for 
logistic regression were obtained by combining L-BFGS  with  carefully designed distributed computations.
\item  
Another stochastic (L-)BFGS method has been proposed to encourage the {\em self-correcting property} of BFGS-type updating. 
In particular, it is known that (L-)BFGS behaves well when the inner product $y_k^Ts_k$ remains bounded above and below (away from zero).  Hence, in \cite{Curt2016}, a modification of the secant equation is proposed to ensure such bounds on $y_k^Ts_k$.  Letting $y_k = \nabla F_{S_{k+1}}(w_{k+1}) - \nabla F_{S_k}(w_k)$ and $s_k=w_{k+1}-w_k$, the method replaces the secant equation with $s_k = B^{-1}v_k$, where $v_k = \beta_k s_k+(1-\beta_k)y_k$ for some $\beta_k\in (0,1)$.  In particular, $\beta_k$ is chosen specifically to ensure that $s_k^Tv_k = \beta_k s_k^T s_k + (1-\beta_k) y_k^T s_k$ is bounded above and below by positive constants.  With this \emph{damping} of the update vectors, the method exhibits stable performance in convex and nonconvex settings.

\item An alternative idea aimed at ensuring that $s_k^Ty_k>0$ has been proposed in \cite{ByrdNoceHansSing:2015} and extended in \cite{GoldGoweRich2016} to a block version of the BFGS method. The key idea is to set $y_k$ not as the difference of stochastic gradient estimates, but as $y_k=\nabla^2 F_{S_k^\prime}(w_k)s_k$ where $\nabla^2 F_{S_k^\prime}(w_k)$ is a sample Hessian computed on some batch $S_k^\prime$, different from the batch that is used to compute the step $s_k$.  This approach has been successful
on convex models such as logistic regression.  However, if the sample Hessian is computed using small mini-batches (relative to $d_w$), then 
$y_k^T s_k=s_k^T\nabla^2 F_{S_k^\prime}(w_k)s_k$ may be small and the method may become unstable. 
\end{itemize}

Stochastic optimization methods (without variance reduction) cannot achieve a convergence rate that is faster than sublinear, even if second-order information is used.  However, using second-order information is an attractive idea since, if the Hessian approximation matrices converge to the Hessian at the solution, then the constant in the convergence rate can be lessened as the effects of ill-conditioning can be reduced.

Unfortunately, despite practical improvements that have been observed, there has yet to be a practical second-order method that provably achieves such  improvements in theory.  As of now, most practical methods only provably achieve the convergence (rate) properties of SGD as long as the Hessian (approximation) matrices remain well-behaved.  For example, if the sequence $\{B_k\}$ (not necessarily produced by BFGS updates) satisfies, for all $k$,
\begin{equation*}
  \begin{aligned}
    \nabla F(w_k)^T \E[B_k^{-1}\nabla_{S_k} F(w_k)] &\geq \mu_1\|\nabla F(w_k)\|^2\ \ \text{for}\ \ \mu_1 \in (0,\infty), \\
    \text{and}\ \ \|\E[B_k^{-1}\nabla_{S_k} F(w_k)]\|^2 &\leq \mu_2\|\nabla F(w_k)\|^2\ \ \text{for}\ \ \mu_2 \in [\mu_1,\infty),
  \end{aligned}
\end{equation*}
then $w_{k+1} \gets w_k - \alpha_k B_k^{-1}\nabla_{S_k} F(w_k)$ achieves the same convergence rate properties as SGD.  It is reasonable to assume that such bounds hold for the method discussed above, with appropriate safeguards; however, one should be careful in a quasi-Newton context in which the stochastic gradient estimates might be correlated with the Hessian approximations.


\section{Deep Learning}\label{sec.dnn}

We have seen that mathematical optimization plays an integral role in machine learning.  In particular, in \S\ref{sec.learning_problem}, we saw that in hopes of solving a given learning problem, a careful sequence of choices and practical approximations must be made before one arrives at a tractable problem such as \eqref{eq:empirical_risk}.  We have also seen how, as in many traditional machine learning models, one can arrive at a convex optimization problem such as~\eqref{eq:minempreglogloss} for which numerous types of algorithms can be applied with well-known strong convergence guarantees.  However, for a variety of types of learning problems, researchers have found that ending up at a convex problem goes perhaps too far in terms of simplifications of the function one is trying to learn to make accurate predictions.  Indeed, recent advances have shown that such convex models are unable to provide predictors that are as powerful as those that can be achieved through more complex \emph{nonconvex} models.  In this section, we provide an overview of recent developments in this area of optimization for machine learning.

The major advancements that have been made along these lines involve the use of \emph{deep neural networks} (DNNs).  The corresponding branch of ML known as \emph{deep learning} (or \emph{hierarchical learning}) represents classes of algorithms that attempt to construct high-level abstractions in data by using a deep graph with multiple \emph{layers} involving sequential linear and nonlinear transformations \cite{bengio2009learning,lecun2015deep,schmidhuber2015deep,hagan1996neural,haykin1994neural,deng2014deep}.  A variety of neural network types have been studied in recent years, including fully-connected networks (FNNs) \cite{tsomokos2008fully,gent1992predicting}, convolutional networks (CNNs) \cite{lecun1995convolutional}, and recurrent networks (RNNs) \cite{HochreiterS97,mikolov2010recurrent,lukovsevivcius2009reservoir}.  For our purposes, we will mainly refer to the first two types of networks, while keeping in mind the others.

While the idea of ``high-level abstractions'' might sound complicated, understanding the basics of DNNs is not that difficult once one is familiar with the terminology.  Deep neural networks derive their name from being inspired by simplifications of models in neuroscience.  It is not necessary to be an expert in neuroscience to understand DNNs, but many do find it convenient to borrow its terminology when describing the structure of a DNN and how one combines input information through a sequence of stages until an ultimate output is produced.  The idea makes sense as to how the human brain takes various pieces of input from multiple sources and  combines them all through a multi-step process (learned since birth) in order to make a prediction.  For example, to classify an image as one of a dog, one combines bits of information---e.g., sections that look like ears, a tail, and fur all oriented in the appropriate manner---to make such a determination.

\subsection{Formulation}

\begin{figure}[ht]
  \centering
  \begin{tikzpicture}[scale=1.0]
  \coordinate[label={[rotate=90]left:Input Layer}]  (inlay) at (-0.8,1.0);
  \coordinate[label={[rotate=270]left:Output Layer}]  (outlay) at (6.8,-1.0);
  \coordinate[label={below:Hidden Layers}]  (hidlay) at (3.0,-2.2);
  \coordinate[label={[label distance=2pt]left:$x_5$}] (l11) at (0.0,-2.0);
  \coordinate[label={[label distance=2pt]left:$x_4$}] (l12) at (0.0,-1.0);
  \coordinate[label={[label distance=2pt]left:$x_3$}] (l13) at (0.0, 0.0);
  \coordinate[label={[label distance=2pt]left:$x_2$}] (l14) at (0.0, 1.0);
  \coordinate[label={[label distance=2pt]left:$x_1$}] (l15) at (0.0, 2.0);
  \coordinate[label={[label distance=3pt]below:$h_{14}$}] (l21) at (2.0,-1.5);
  \coordinate[label={[label distance=3pt]below:$h_{13}$}] (l22) at (2.0,-0.5);
  \coordinate[label={[label distance=3pt]above:$h_{12}$}] (l23) at (2.0, 0.5);
  \coordinate[label={[label distance=3pt]above:$h_{11}$}] (l24) at (2.0, 1.5);
  \coordinate[label={[label distance=3pt]below:$h_{24}$}] (l31) at (4.0,-1.5);
  \coordinate[label={[label distance=3pt]below:$h_{23}$}] (l32) at (4.0,-0.5);
  \coordinate[label={[label distance=3pt]above:$h_{22}$}] (l33) at (4.0, 0.5);
  \coordinate[label={[label distance=3pt]above:$h_{21}$}] (l34) at (4.0, 1.5);
  \coordinate[label={[label distance=2pt]right:$p_3$}] (l41) at (6.0,-1.0);
  \coordinate[label={[label distance=2pt]right:$p_2$}] (l42) at (6.0, 0.0);
  \coordinate[label={[label distance=2pt]right:$p_1$}] (l43) at (6.0, 1.0);
  \draw[green,fill=white] (l11) circle (4pt);
  \draw[green,fill=white] (l12) circle (4pt);
  \draw[green,fill=white] (l13) circle (4pt);
  \draw[green,fill=white] (l14) circle (4pt);
  \draw[green,fill=white] (l15) circle (4pt);
  \draw[blue,fill=white] (l21) circle (4pt);
  \draw[blue,fill=white] (l22) circle (4pt);
  \draw[blue,fill=white] (l23) circle (4pt);
  \draw[blue,fill=white] (l24) circle (4pt);
  \draw[blue,fill=white] (l31) circle (4pt);
  \draw[blue,fill=white] (l32) circle (4pt);
  \draw[blue,fill=white] (l33) circle (4pt);
  \draw[blue,fill=white] (l34) circle (4pt);
  \draw[red,fill=white] (l41) circle (4pt);
  \draw[red,fill=white] (l42) circle (4pt);
  \draw[red,fill=white] (l43) circle (4pt);
  \def\ptRad{6pt}
  \draw[->, shorten <=\ptRad,shorten >=\ptRad] (l11) -- (l21); 
  \draw[->, shorten <=\ptRad,shorten >=\ptRad] (l11) -- (l22); 
  \draw[->, shorten <=\ptRad,shorten >=\ptRad] (l11) -- (l23); 
  \draw[->, shorten <=\ptRad,shorten >=\ptRad] (l11) -- (l24); 
  \draw[->, shorten <=\ptRad,shorten >=\ptRad] (l12) -- (l21); 
  \draw[->, shorten <=\ptRad,shorten >=\ptRad] (l12) -- (l22); 
  \draw[->, shorten <=\ptRad,shorten >=\ptRad] (l12) -- (l23); 
  \draw[->, shorten <=\ptRad,shorten >=\ptRad] (l12) -- (l24); 
  \draw[->, shorten <=\ptRad,shorten >=\ptRad] (l13) -- (l21); 
  \draw[->, shorten <=\ptRad,shorten >=\ptRad] (l13) -- (l22); 
  \draw[->, shorten <=\ptRad,shorten >=\ptRad] (l13) -- (l23); 
  \draw[->, shorten <=\ptRad,shorten >=\ptRad] (l13) -- (l24); 
  \draw[->, shorten <=\ptRad,shorten >=\ptRad] (l14) -- (l21); 
  \draw[->, shorten <=\ptRad,shorten >=\ptRad] (l14) -- (l22); 
  \draw[->, shorten <=\ptRad,shorten >=\ptRad] (l14) -- (l23); 
  \draw[->, shorten <=\ptRad,shorten >=\ptRad] (l14) -- (l24); 
  \draw[->, shorten <=\ptRad,shorten >=\ptRad] (l15) -- (l21); 
  \draw[->, shorten <=\ptRad,shorten >=\ptRad] (l15) -- (l22); 
  \draw[->, shorten <=\ptRad,shorten >=\ptRad] (l15) -- (l23); 
  \draw[->, shorten <=\ptRad,shorten >=\ptRad] (l15) -- (l24); 
  \draw[->, shorten <=\ptRad,shorten >=\ptRad] (l21) -- (l31); 
  \draw[->, shorten <=\ptRad,shorten >=\ptRad] (l21) -- (l32); 
  \draw[->, shorten <=\ptRad,shorten >=\ptRad] (l21) -- (l33); 
  \draw[->, shorten <=\ptRad,shorten >=\ptRad] (l21) -- (l34); 
  \draw[->, shorten <=\ptRad,shorten >=\ptRad] (l22) -- (l31); 
  \draw[->, shorten <=\ptRad,shorten >=\ptRad] (l22) -- (l32); 
  \draw[->, shorten <=\ptRad,shorten >=\ptRad] (l22) -- (l33); 
  \draw[->, shorten <=\ptRad,shorten >=\ptRad] (l22) -- (l34); 
  \draw[->, shorten <=\ptRad,shorten >=\ptRad] (l23) -- (l31); 
  \draw[->, shorten <=\ptRad,shorten >=\ptRad] (l23) -- (l32); 
  \draw[->, shorten <=\ptRad,shorten >=\ptRad] (l23) -- (l33); 
  \draw[->, shorten <=\ptRad,shorten >=\ptRad] (l23) -- (l34); 
  \draw[->, shorten <=\ptRad,shorten >=\ptRad] (l24) -- (l31); 
  \draw[->, shorten <=\ptRad,shorten >=\ptRad] (l24) -- (l32); 
  \draw[->, shorten <=\ptRad,shorten >=\ptRad] (l24) -- (l33); 
  \draw[->, shorten <=\ptRad,shorten >=\ptRad] (l24) -- (l34); 
  \draw[->, shorten <=\ptRad,shorten >=\ptRad] (l31) -- (l41); 
  \draw[->, shorten <=\ptRad,shorten >=\ptRad] (l31) -- (l42); 
  \draw[->, shorten <=\ptRad,shorten >=\ptRad] (l31) -- (l43); 
  \draw[->, shorten <=\ptRad,shorten >=\ptRad] (l32) -- (l41); 
  \draw[->, shorten <=\ptRad,shorten >=\ptRad] (l32) -- (l42); 
  \draw[->, shorten <=\ptRad,shorten >=\ptRad] (l32) -- (l43); 
  \draw[->, shorten <=\ptRad,shorten >=\ptRad] (l33) -- (l41); 
  \draw[->, shorten <=\ptRad,shorten >=\ptRad] (l33) -- (l42); 
  \draw[->, shorten <=\ptRad,shorten >=\ptRad] (l33) -- (l43); 
  \draw[->, shorten <=\ptRad,shorten >=\ptRad] (l34) -- (l41); 
  \draw[->, shorten <=\ptRad,shorten >=\ptRad] (l34) -- (l42); 
  \draw[->, shorten <=\ptRad,shorten >=\ptRad] (l34) -- (l43); 
  \coordinate[label={[label distance=3pt]above:$[W_1]_{54}$}] (w1) at (1.0,-2.5);
  \coordinate[label={[label distance=3pt]below:$[W_1]_{11}$}] (w2) at (1.0, 2.5);
  \coordinate[label={[label distance=3pt]above:$[W_2]_{44}$}] (w3) at (3.0,-2.2);
  \coordinate[label={[label distance=3pt]below:$[W_2]_{11}$}] (w4) at (3.0, 2.2);
  \coordinate[label={[label distance=3pt]above:$[W_3]_{43}$}] (w5) at (5.0,-2.0);
  \coordinate[label={[label distance=3pt]below:$[W_3]_{11}$}] (w6) at (5.0, 2.0);
\end{tikzpicture}
  \caption{Small fully-connected neural network with 2 hidden layers.}
  \label{fig: toy_dnn}
\end{figure}
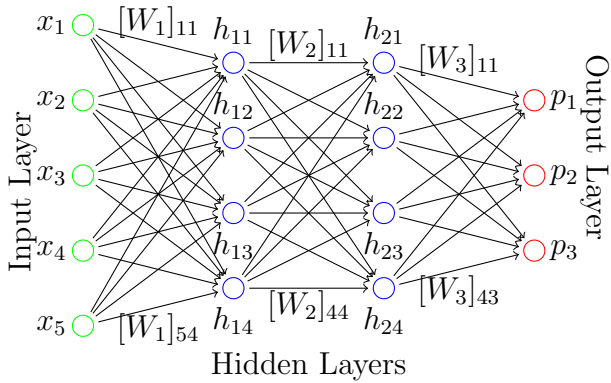

Structurally, a DNN takes the form of a graph with subsets of nodes arranged in a sequence.  Each subset of nodes (or \emph{neurons}) is called a \emph{layer}, where in simple cases edges only exist between neurons in a layer and the subsequent layer.  However, besides its structure, the key aspect of a DNN is how information is ``fed'' through it.  In the simple case of a \emph{feed forward} network, this occurs as follows.  First, each element of an input vector $x = (x_1,\dots,x_{d_x})$ is given to a different neuron in the first layer, also known as the \emph{input layer}.  The values in this layer are then each passed to the neurons in the next layer after multiplication with \emph{weights} associated with the corresponding edges.  Once at a given node, a value can be further transformed through application of a (linear or nonlinear) \emph{activation} function before values continue to be passed through the network.  The last layer of the network, which provides the predicted output $p(x)$, is known as the \emph{output layer}, whereas layers between the input and output are called \emph{hidden layers}.  The more hidden layers that are present, the deeper the network.  Figure \ref{fig: toy_dnn} provides a partial illustration of a small fully-connected DNN for a simple classification model with $d_x=5$, $d_y=3$, and two hidden layers.

Mathematically, the application of a DNN starts by setting $x^{(1)} \gets x$ as the input layer values.  One then applies repeated transformations (for each subsequent layer) of the form
\begin{equation}\label{eq.layer_formula}
  x^{(j+1)} \gets s_j(W_jx^{(j)} + \omega_j)
\end{equation}
where each $W_j$ is a matrix of edge weight parameters, $\omega_j$ is a vector of shift parameters, and $s_j$ is an activation function (common choices of which are componentwise sigmoid, tanh, and ReLU (rectified linear unit) \cite{dahl2013improving}, along with the latter's variants such as PReLU (Parametric-ReLU) \cite{maas2013rectifier} and LReLU (Leaky-ReLU) \cite{he2015delving}). For example, in the case of the network depicted in Figure~\ref{fig: toy_dnn}, $h_{11}$ is the first component of the vector $s_1(W_1x^{(1)})$ while $h_{23}$ is the third component of the vector
$s_2(W_2h_1)$, where $h_1=(h_{11}, h_{12}, h_{13}, h_{14})$.  The parameters of this network are the matrices $W_1\in\R^{4\times 5}$, $W_2\in\R^{4\times 4}$, and $W_3\in\R^{3\times 4}$, as well as the vectors $\omega_1\in \R^4$, $\omega_2\in \R^4$, and $\omega_3\in \R^3$.  Thus, in total, this network has $59$ parameters.

Other graph structures are common in deep learning, all designed, to some extent, for specific purposes.  For example, {\em convolutional neural networks} (CNNs) have received a lot of attention in the context of image recognition (and other perceptual tasks).  These networks exploit the fact that the input consists of units, such as pixels of an image, that are inherently connected due to their spatial relationships.

\begin{figure}[ht]
  \centering
  \scalebox{0.75}{\begin{tikzpicture}[scale=1.0]
  \draw[white,pattern=north west lines,pattern color=gray!40] (5,2) -- (7,2) -- (7,4) -- (5,4) -- (5,2);
  \draw[white,pattern=north west lines,pattern color=gray!40] (10,2.5) -- (11,2.5) -- (11,3.5) -- (10,3.5) -- (10,2.5);
  \draw[white,pattern=north east lines,pattern color=gray!40] (6,0) -- (8,0) -- (8,2) -- (6,2) -- (6,0);
  \draw[white,pattern=north east lines,pattern color=gray!40] (11,0.5) -- (12,0.5) -- (12,1.5) -- (11,1.5) -- (11,0.5);
  \draw[black] (0,0) -- (4,0);
  \draw[black] (0,1) -- (4,1);
  \draw[black] (0,2) -- (4,2);
  \draw[black] (0,3) -- (4,3);
  \draw[black] (0,4) -- (4,4);
  \draw[black] (0,0) -- (0,4);
  \draw[black] (1,0) -- (1,4);
  \draw[black] (2,0) -- (2,4);
  \draw[black] (3,0) -- (3,4);
  \draw[black] (4,0) -- (4,4);
  \coordinate[label={[label distance=11pt]above right:$1$}] (d11) at (0,0);
  \coordinate[label={[label distance=11pt]above right:$0$}] (d12) at (1,0);
  \coordinate[label={[label distance=11pt]above right:$9$}] (d13) at (2,0);
  \coordinate[label={[label distance=11pt]above right:$2$}] (d14) at (3,0);
  \coordinate[label={[label distance=11pt]above right:$2$}] (d21) at (0,1);
  \coordinate[label={[label distance=11pt]above right:$8$}] (d22) at (1,1);
  \coordinate[label={[label distance=11pt]above right:$0$}] (d23) at (2,1);
  \coordinate[label={[label distance=11pt]above right:$8$}] (d24) at (3,1);
  \coordinate[label={[label distance=11pt]above right:$9$}] (d31) at (0,2);
  \coordinate[label={[label distance=11pt]above right:$1$}] (d32) at (1,2);
  \coordinate[label={[label distance=11pt]above right:$7$}] (d33) at (2,2);
  \coordinate[label={[label distance=11pt]above right:$0$}] (d34) at (3,2);
  \coordinate[label={[label distance=11pt]above right:$1$}] (d41) at (0,3);
  \coordinate[label={[label distance=11pt]above right:$8$}] (d42) at (1,3);
  \coordinate[label={[label distance=11pt]above right:$0$}] (d43) at (2,3);
  \coordinate[label={[label distance=11pt]above right:$2$}] (d44) at (3,3);
  \draw[black,very thick] (4.5,0) -- (4.5,4);
  \draw[black] (5,0) -- (9,0);
  \draw[black] (5,1) -- (9,1);
  \draw[black] (5,2) -- (9,2);
  \draw[black] (5,3) -- (9,3);
  \draw[black] (5,4) -- (9,4);
  \draw[black] (5,0) -- (5,4);
  \draw[black] (6,0) -- (6,4);
  \draw[black] (7,0) -- (7,4);
  \draw[black] (8,0) -- (8,4);
  \draw[black] (9,0) -- (9,4);
  \coordinate[label={[label distance=11pt]above right:$1$}] (D11) at (5,0);
  \coordinate[label={[label distance=11pt]above right:$0$}] (D12) at (6,0);
  \coordinate[label={[label distance=11pt]above right:$9$}] (D13) at (7,0);
  \coordinate[label={[label distance=11pt]above right:$2$}] (D14) at (8,0);
  \coordinate[label={[label distance=11pt]above right:$2$}] (D21) at (5,1);
  \coordinate[label={[label distance=11pt]above right:$8$}] (D22) at (6,1);
  \coordinate[label={[label distance=11pt]above right:$0$}] (D23) at (7,1);
  \coordinate[label={[label distance=11pt]above right:$8$}] (D24) at (8,1);
  \coordinate[label={[label distance=11pt]above right:$9$}] (D31) at (5,2);
  \coordinate[label={[label distance=11pt]above right:$1$}] (D32) at (6,2);
  \coordinate[label={[label distance=11pt]above right:$7$}] (D33) at (7,2);
  \coordinate[label={[label distance=11pt]above right:$0$}] (D34) at (8,2);
  \coordinate[label={[label distance=11pt]above right:$1$}] (D41) at (5,3);
  \coordinate[label={[label distance=11pt]above right:$8$}] (D42) at (6,3);
  \coordinate[label={[label distance=11pt]above right:$0$}] (D43) at (7,3);
  \coordinate[label={[label distance=11pt]above right:$2$}] (D44) at (8,3);
  \draw[black] (10,0.5) -- (13,0.5);
  \draw[black] (10,1.5) -- (13,1.5);
  \draw[black] (10,2.5) -- (13,2.5);
  \draw[black] (10,3.5) -- (13,3.5);
  \draw[black] (10,0.5) -- (10,3.5);
  \draw[black] (11,0.5) -- (11,3.5);
  \draw[black] (12,0.5) -- (12,3.5);
  \draw[black] (13,0.5) -- (13,3.5);
  \coordinate[label={[label distance=11pt]above right: $9$}] (e11) at (10,0.5);
  \coordinate[label={[label distance=11pt]above right: $0$}] (e12) at (11,0.5);
  \coordinate[label={[label distance=11pt]above right:$17$}] (e13) at (11.9,0.5);
  \coordinate[label={[label distance=11pt]above right: $3$}] (e21) at (10,1.5);
  \coordinate[label={[label distance=11pt]above right:$15$}] (e22) at (10.9,1.5);
  \coordinate[label={[label distance=11pt]above right: $0$}] (e23) at (12,1.5);
  \coordinate[label={[label distance=11pt]above right:$17$}] (e31) at (9.9,2.5);
  \coordinate[label={[label distance=11pt]above right: $1$}] (e32) at (11,2.5);
  \coordinate[label={[label distance=11pt]above right: $9$}] (e33) at (12,2.5);
  \draw [red,thick,rounded corners,-latex] (6.5,4) -- (6.5,4.25) -- (10.5,4.25) -- (10.5,3.5);
  \draw [red,thick,rounded corners,-latex] (7.5,0) -- (7.5,-0.25) -- (11.5,-0.25) -- (11.5,0.5);
  \draw[black] (10.75,3.75) -- (11.75,3.75);
  \draw[black] (10.75,4.25) -- (11.75,4.25);
  \draw[black] (10.75,4.75) -- (11.75,4.75);
  \draw[black] (10.75,3.75) -- (10.75,4.75);
  \draw[black] (11.25,3.75) -- (11.25,4.75);
  \draw[black] (11.75,3.75) -- (11.75,4.75);
  \coordinate[label={[label distance=1.5pt]above right:$1$}] (f11) at (10.75,3.75);
  \coordinate[label={[label distance=1.5pt]above right:$0$}] (f12) at (11.25,3.75);
  \coordinate[label={[label distance=1.5pt]above right:$0$}] (f21) at (10.75,4.25);
  \coordinate[label={[label distance=1.5pt]above right:$1$}] (f22) at (11.25,4.25);
\end{tikzpicture}}
  \caption{Application of a filter in a convolutional layer.  The original $4 \times 4$ data appears at the left.  On the right is illustrated how the filter---i.e., the $2 \times 2$ matrix on the top left---can be passed over the data matrix to produce the $3 \times 3$ feature map at the right.}
  \label{fig.cnn}
\end{figure}
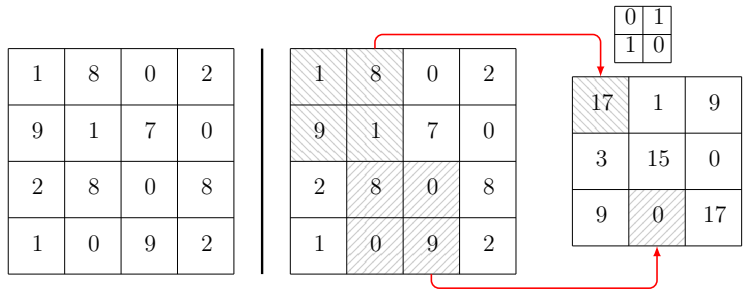

Let us describe some of the ideas underlying CNNs by describing the operation of its key building block, a \emph{convolutional layer}, first with a numerical example and then more broadly.  Consider the matrix of data values on the left-hand side of Figure~\ref{fig.cnn}.  If one has a smaller matrix, call it a \emph{filter}, then the application of a convolutional layer results in another smaller matrix formed by ``sliding'' the filter across the original matrix, taking the dot product of each overlap, and storing the results.  For example, in Figure~\ref{fig.cnn}, we illustrate the application of a $2\times2$ filter to $4\times4$ data, resulting in a $3\times3$ matrix.  The parameters to be optimized in this convolutional layer are the entries of the filter (of which there might be many in a particular layer, resulting in a set of smaller matrices, rather than only one).  Even though we illustrate this example with matrices, one can easily translate these operations to a graph structure where each value in the input data corresponds to a node in the input layer, then a \emph{common} weight matrix (and, potentially, a common shift) is applied to obtain the values in the smaller matrix corresponding to values in the first hidden layer.

What role can a filter play and how might it help with image recognition?  Suppose that the data matrix in Figure~\ref{fig.cnn} represents the grayscale values of a two-dimensional image, with larger numbers corresponding to brighter pixels and smaller numbers corresponding to darker ones.  Then, a filter such as the one illustrated might help to detect sections with a bright diagonal pattern.  Extrapolating this idea to larger images and more complicated inputs (say, with RGB values), one can imagine optimizing a set of filters, each intended to detect different patterns, such as edges or blobs of color.  Each matrix produced by applying a filter is referred to as a \emph{feature map}, which can be viewed as an image itself in which all patterns have been filtered out except the one of interest for that filter.  It is worthwhile to mention that the idea of creating feature maps was inspired by processes in neuroscience---in this case, the connectivity between neurons in an animal's visual cortex.

\begin{figure}[ht]
    \centering
    \includegraphics[width=0.8\textwidth]{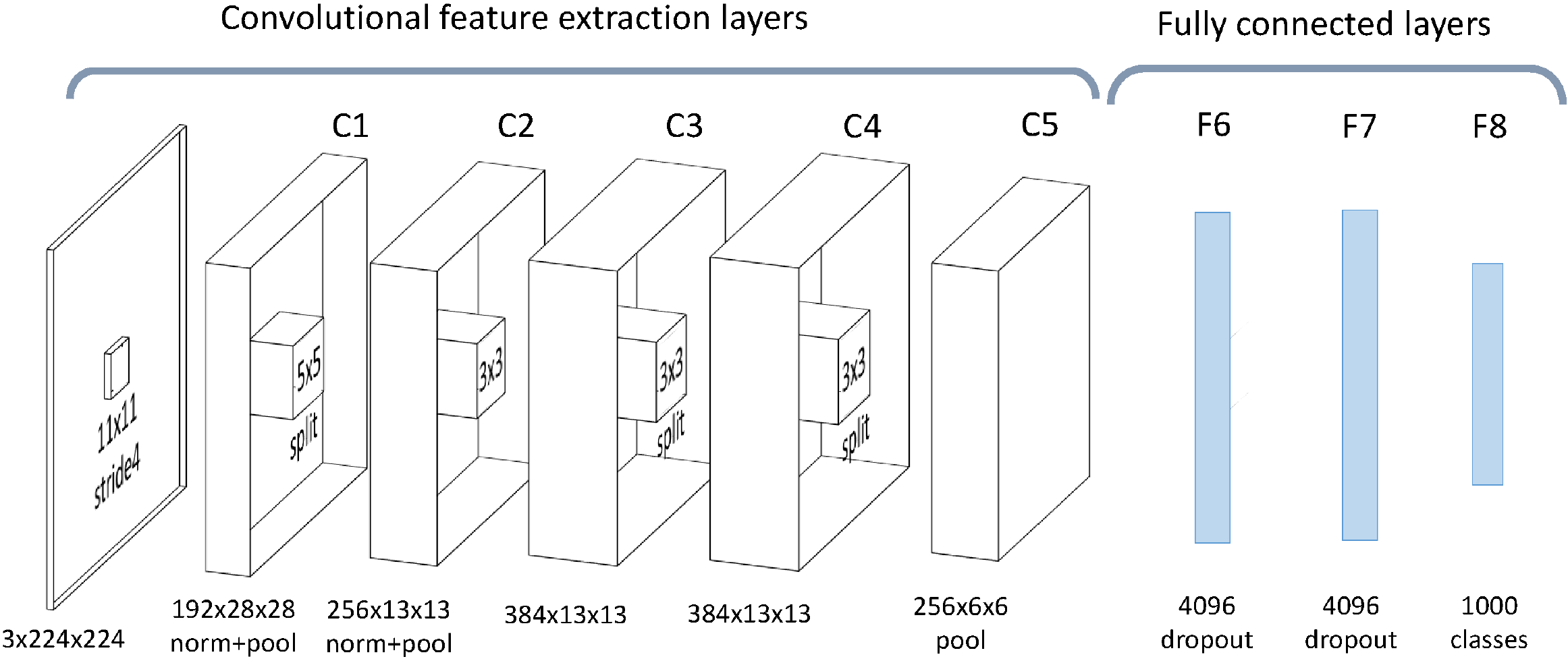}
    \caption{CNN for image classification applied to the ILSVRC challenge \cite{BottCurtNoce16}.}
    \label{fig: ilsvrc}
\end{figure}

CNNs involve other types of building blocks as well (e.g., pooling layers), all arranged carefully in sequence.  We leave the reader to explore the literature for further information.  For our purposes, let us, as an example, simply show the structure of a complete CNN; see Figure~\ref{fig: ilsvrc}.  This CNN improved the state-of-the-art of image classification tasks for the ImageNet Large Scale Visual Recognition Challenge \cite{ILSVRC15}.  In this challenge, the dataset contained around 1.2 million examples.  The network involves around $60$ million parameters.

For DNNs in general, the structure of the graph and the activation functions are typically determined in advance, meaning that, to \emph{train it}, one needs to solve an optimization problem to minimize expected (or empirical) risk over the parameters $w = (W_1,\omega_1,W_2,\omega_2,\dots)$.  Unfortunately, however, such a problem cannot in general be reduced to a convex optimization problem.  Indeed, the resulting problems are highly nonconvex with numerous saddle points and local minimizers.  Due to this, one might expect that training a DNN would be an impossible task.  However, with the vast amounts of data and high-performance computing power available today, researchers have found that optimization methods can offer (approximate) solutions leading to DNNs with great predictive powers.

The increased popularity of neural networks and deep learning in recent years has been due to both theoretical and practical reasons. The universal approximation theorem established in $1989$ \cite{hornik1989multilayer} shows that a feed-forward neural network with only a single hidden layer could, under mild assumptions on the activation function, approximate any continuous function on a compact subset of $\R^d$.  On the practical side, recent advances in optimization techniques and the use of computational resources have helped to train DNNs for large-scale applications, such as for speech recognition \cite{hinton2012deep,graves2013speech}, image classification \cite{ciregan2012multi,krizhevsky2012imagenet,simonyan2013deep}, human action recognition \cite{ji20133d}, video classification \cite{karpathy2014large}, forecasting \cite{deo2001neural,guo2012multi,thirumalaiah1998river,tsai1999back}, machine translation \cite{cho2014learning,bahdanau2014neural}, and civil engineering \cite{flood1994neural,adeli2001neural,kartam1997artificial,elkordy1993neural}; see also~\cite{yu2011deep}.

\subsection{Stochastic gradient descent}

Given the complex structure of DNNs, one might think it would be difficult to implement an optimization algorithm to optimize  the parameter vector $w$ in a problem of the form~\eqref{eq:empirical_risk} when $p(w,x_i)$ comes from a series of evaluations of the form~\eqref{eq.layer_formula}.  In particular, how can (stochastic) gradients be computed?  Fortunately, in the 1980s, a key observation was made that such gradients can be computed efficiently using a process known as \emph{back propagation} \cite{backprop,lecun-98b}.  In essence, this involves the application of the chain rule through automatic differentiation.   For this tutorial, it is not necessary to look into the details of back propagation, but it is important to recognize its importance in being able to efficiently apply algorithms such as (stochastic) gradient descent and its variants, e.g., that we have described in \S\ref{sec:Logreg}.

But what can be said about convergence guarantees for such methods?  After all, the optimization problems in~\S\ref{sec:Logreg} are convex or even strongly convex, a setting in which nice theoretical guarantees can be provided.  What about training DNNs when the objective function is nonconvex with numerous local minimizers and saddle points?  Therein lies a number of fascinating questions about the use of DNNs that will inevitably lead to new research directions for years to come.  In short, while some theoretical guarantees (both for optimization and for learning) have been provided, the practical gains achieved by the use of DNNs in recent years remains somewhat of a mystery.

Let us highlight the enigmatic behavior of optimization algorithms applied to train DNNs by citing the following.  First, one has, e.g., in~\cite{BottCurtNoce16}, a result which says that by applying SGD to minimize a nonconvex objective function (drawing indefinitely from the input $\times$ output space), one has a guarantee that the gradient of the expected risk will vanish, at least over a subsequence; i.e., $ \lim\inf_{k\to\infty} \E[\|\nabla f(x_k)\|_2] = 0$.  This result  is comforting in that it shows that SGD can achieve a convergence guarantee similar to other state-of-the-art gradient-based optimization algorithms.  However, this and other guarantees in the literature are limited; after all, whereas many gradient-based optimization algorithms ensure monotonic decrease of the objective function, SG does not operate in this manner.  So if a subsequence is converging to a stationary point, how can we be sure that the stationary point is not a saddle point, poor local minimizer, or perhaps some maximizer with objective value worse than the initial point?  In truth, we cannot be sure.  That said, SGD does often seem to find good local, if not global minima. On the other hand, it tends to slow down around stationary points, which can hinder its progress in training DNNs \cite{negativecurvatures, PascanuDGB14,Bengiononconvex}.

In general, for nonconvex problems, convergence rates results for SGD do exist \cite{ghadimilannonconvex1,ghadimilannonconvex2}, but they are very limited, and in particular they are not applicable to the discussion in \S\ref{sec:learningbounds}.  Hence, we cannot argue in the same manner that SGD is a superior method for nonconvex optimization problems in ML.  Moreover, learning bounds such as that in \eqref{eq:estimerr} are not useful, because for many DNNs and CNNs the complexity $C$ of the classifier class produced by a neural network is much bigger than the number of training examples $n$.  In fact, in \cite{RechHardt} it is empirically shown that neural networks can easily overfit typical types of datasets, if only the data in these sets are randomly perturbed.  The information inherent in a typical ``real'' dataset is ruined by random perturbations, yet that does not stop the training process for a DNN (e.g., using the momentum variant of SGD described in Algorithm \ref{alg:msgd}) to find a ``good'' solution for a dataset that is actually completely meaningless!

It remains an open question how one can design stochastic-gradient-type methods to optimize parameters of a DNN so as to find good local minimizers and avoid poor local minimizers and/or saddle points.  Stochastic quasi-Newton methods described in \S\ref{sec:bfgs} have been applied to DNNs with mixed results. However, clearly, they are not aimed at exploiting negative curvature, since they construct positive definite Hessian approximations.  One alternative avenue that is being explored in the context of deep learning is the use of non-stochastic second-order methods, which have the benefit of observing accurate curvature information about the objective function, so as to hopefully avoid saddle points.

\subsection{Hessian-free methods}\label{sec:HF}

Let us now briefly describe a class of methods that form $B_k$ in \eqref{eq:quadmod} as the true or a modified Hessian matrix using the entire dataset.  While such an approach might sound prohibitively expensive, it is possible to employ such an approach through efficient use of \emph{parallel and distributed computation} across a network of computing nodes.  In such settings, it is possible to use full batch computation (e.g., to compute gradients exactly), though it is still not reasonable to assume that one can compute \emph{and store} a complete Hessian matrix.  Fortunately, however, in certain Newton-type methods one only needs a subroutine for computing Hessian-vector products, i.e., products of a Hessian with a set of vectors.  (For example, this is the case in \emph{Newton-CG}, wherein a step is computed by minimizing \eqref{eq:quadmod} approximately using the conjugate gradient algorithm.)  For these purposes, one finds that the back propagation algorithm for DNNs can be modified to compute such Hessian-vector products, since they can be seen as directional derivatives \cite{pearlmutter}.  The cost of computing such a product is only a constant factor more than computing a gradient.  The resulting class of methods is often referred to as \emph{Hessian-free} since, while Hessian information is accessed and used, no Hessian matrix is ever stored explicitly.


Additional complications arise in the context of DNNs since, due to nonconvexity of the objective function, the true Hessian matrix might not be positive definite.  In general, as in deterministic optimization, two possible ways of handling this issue are to \emph{modify} the Hessians or employ a \emph{trust region} methodology.  Both of these directions are being explored in the context of training DNNs.  For example, in \cite{martenshfpaper,martenschapter}, 
a Gauss-Newton method is proposed, which approximates the Hessian matrix by the first term in the following formula for the Hessian of $F$ in \eqref{eq:P} (ignoring the regularization term):
\begin{equation*}
  \nabla^2 F(w) = \frac{1}{n} \sum_{i=1}^n \left [\nabla p(w,x_i)^T \nabla^2 \ell(p(w,x_i),y_i) \nabla p(w,x_i) + \sum_{j=1}^{d_y} [\ell(p_j(w,x_i),y_i)\nabla^2 {[p_j(w,x_i)]}\right],
\end{equation*}
where $\nabla^2 \ell(p(w,x_i),y_i)$ is the Hessian of the loss function $\ell(\cdot,\cdot)$ with respect to the first argument, $\nabla p(w,x_i)$  is the Jacobian of the $d_y$-dimensional function $p(w,x)$ with respect to $w$, and $\nabla^2 {[p_j(w,x_i)]}$ for all $j \in \{1,\dots,d_y\}$ are the element-wise Hessians with respect to $w$.  The sum of the first terms, known as the Gauss-Newton matrix, is positive semidefinite.  If the algorithm converges to a solution with zero loss, then $ \ell(p_j(w,x_i),z_i) \rightarrow 0$, meaning that the Gauss-Newton matrix converges to the true Hessian in the limit, which in turn must be positive semidefinite.  However, when there is nonzero loss, the Gauss-Newton matrix differs from the true Hessian and the method may converge to a saddle point.  Since the Gauss-Newton matrix is not necessarily nonsingular, a small regularization term of the form $\gamma I$ is often added when forming $B_k$.  In any case, to obtain $s_k \approx -B_k^{-1}\nabla F(w_k)$, the methods in \cite{martenshfpaper,martenschapter} use a (preconditioned) conjugate gradient method.
 
The idea of using a trust region methodology, say with $B_k$ representing the true Hessian, holds a lot of promise.  That said, to maintain a Hessian-free nature, the trust region subproblem must be solved (approximately) using only Hessian-vector products. Two well-known algorithms that serve this purpose are the Steihaug-Toint CG method~\cite{steihaug,toint} and GLTR~\cite{gouldgltr}.  For training DNNs, numerical comparisons of these methods, as well as some newer alternatives, can be found in \cite{Alirezathesis}.  In this study, all methods use 20--30 Hessian-vector multiplications per iteration.  It is found that the methods typically converge to different solutions with different testing losses and accuracies.  One can go further and compute and follow directions of negative curvature (requiring additional Hessian-vector products), which indeed can lead to solutions with better properties.  Additional experimentation and comparisons are needed to demonstrate the potential benefits of such approaches, but in any case it is clear that for such ideas to be effective, the low cost of computing Hessian-vector products (versus computing and storing exact Hessian information) is essential.

\subsection{Subsampled Hessian methods}\label{sec:SH}

Several variants of subsampled Newton methods have been proposed and analyzed for solving (strongly) convex formulations of problem~\eqref{eq:P}
\cite{ RoKhoMah2016I,RoKhoMah2016II,ErdoMont2015, BollByrdNoce2016,Xuetal2016}.  All of these methods  use $B_k = \nabla^2_{S_k}F(w_k)$ as a Hessian approximation, where ${S_k}$ is a random sample set.  Some of these methods also use stochastic gradients, though some assume that gradients are computed exactly.  In all cases, in contrast with the SGD, the size of ${S_k}$  is increased to improve the accuracy of the (gradient and) Hessian estimates as the algorithms converge. The schedule  of the increase of ${S_k}$ and the final convergence results vary depending on the algorithm and the analysis. The methods also differ by how the resulting Newton step is computed.  In general, the convergence rates that are recovered are similar to their deterministic counterparts. 

For strongly convex problems, one can select a step size without performing a line search, obviating the need to evaluate the objective function during the optimization process.  However, for nonconvex problems such as DNN training, evaluating $F$ is essential for ensuring progress.  (For example, in trust region methods, one must accept or reject the step, and increase or decrease the trust region radius, based on whether or not $F$ decreases for a given step.)  On the other hand, as pointed out earlier, each trust-region subproblem solve requires several Hessian-vector products. Hence, using a sampled Hessian for such products can mean significant savings in the overall computational effort per iteration, while evaluating $F$ and possibly even $\nabla F$ is comparatively inexpensive.  These ideas have been used in \cite{martenshfpaper,martenschapter}, though without theoretical justification. 

Recently, in a series of papers \cite{bandeira2013convergence, CartisScheinberg2014, GrattonEtAl2017}, trust region, line search, and adaptive cubic regularization methods have been analyzed in convex and nonconvex cases using a very general framework of random models.  In this work, it is shown that standard optimization methods that use random inexact gradient and Hessian information retain their convergence rates as long as the gradient and Hessian estimates are sufficiently accurate with some positive probability. In the case of machine learning and sampled Hessian and gradients, the results simply require that $|S_k|$ has to be chosen sufficiently large with respect to the length of the step taken by the algorithm.  For example, in \cite{bandeira2013convergence, GrattonEtAl2017}, the size of $|S_k|$ is connected to the trust region radius. It is important to note that the requirement on the size of the sample set for a sampled Hessian is significantly looser than that for the sampled gradient, supporting the idea that using inexpensive Hessian estimates with  exact gradients gives rise to algorithms with strong theoretical behavior and good practical benefits.  In \cite{ChenMenickellyScheinberg2014,  BlanchetCartisMenickellyScheinberg2016}, convergence and convergence rate analyses of a trust region method are carried out under relaxed conditions where $F$ is also computed inexactly, say, by sampling. The requirements on the sample set for evaluating $F$ is even stronger than those for the gradient.  Still, this extension allows trust region methodologies to be employed to minimize stochastic functions such as \eqref{eq:expected_risk}.

 \bibliographystyle{plain}
 \bibliography{OPTML_Review,SODNN_Review,Katya.bib,fec,cxm}

\begin{thebibliography}{10}

\bibitem{adeli2001neural}
H.~Adeli.
\newblock Neural networks in civil engineering: 1989--2000.
\newblock {\em Computer-Aided Civil and Infrastructure Engineering},
  16(2):126--142, 2001.

\bibitem{bahdanau2014neural}
Dzmitry Bahdanau, Kyunghyun Cho, and Yoshua Bengio.
\newblock Neural machine translation by jointly learning to align and
  translate.
\newblock {\em arXiv:1409.0473}, 2014.

\bibitem{bandeira2013convergence}
Afonso~S Bandeira, Katya Scheinberg, and Luis~Nunes Vicente.
\newblock Convergence of trust-region methods based on probabilistic models.
\newblock {\em SIAM Journal on Optimization}, 24(3):1238--1264, 2014.

\bibitem{BartlettMandel}
P.~L. Bartlett and S.~Mendelson.
\newblock Rademacher and gaussian complexities: Risk bounds and structural
  results.
\newblock {\em J. Mach. Learn. Res.}, 3:463--482, March 2003.

\bibitem{Beck-Teboulle-2009}
A.~Beck and M.~Teboulle.
\newblock A fast iterative shrinkage-thresholding algorithm for linear inverse
  problems.
\newblock {\em SIAM J. Imaging Sciences}, 2(1):183--202, 2009.

\bibitem{bengio2009learning}
Y.~Bengio.
\newblock Learning deep architectures for ai.
\newblock {\em Foundations and trends{\textregistered} in Machine Learning},
  2(1):1--127, 2009.

\bibitem{BeraNoceTaka2016}
Albert~S. Berahas, Jorge Nocedal, and Martin Takac.
\newblock A multi-batch l-bfgs method for machine learning.
\newblock In {\em Proceedings of the 29th International Conference on Neural
  Information Processing Systems}, NIPS'16, Cambridge, MA, USA, 2016. MIT
  Press.

\bibitem{BlanchetCartisMenickellyScheinberg2016}
J.~Blanchet, C.~Cartis, M.~Menickelly, and K.~Scheinberg.
\newblock Convergence rate analysis of a stochastic trust region method for
  nonconvex optimization.
\newblock 2016.

\bibitem{BollByrdNoce2016}
R.~Bollapragada, R.~Byrd, and J.~Nocedal.
\newblock Exact and inexact subsampled newton methods for optimization.
\newblock {\em arXiv:1609.08502}, 2016.

\bibitem{Bordesetal}
A.~Bordes, L.~Bottou, and P.~Gallinari.
\newblock Model selection through sparse maximum likelihood estimation for
  multivariate gaussian or binary data.
\newblock {\em Journal of Machine Learning Research}, 10:1737--1754, December
  2009.

\bibitem{BottCurtNoce16}
L.~Bottou, F.~E. Curtis, and J.~Nocedal.
\newblock {Optimization Methods for Large-Scale Machine Learning}.
\newblock Technical Report 1606.04838, arXiv, 2016.

\bibitem{BottouBousquet}
O.~Bousquet and L.~Bottou.
\newblock The tradeoffs of large scale learning.
\newblock In J.~C. Platt, D.~Koller, Y.~Singer, and S.~T. Roweis, editors, {\em
  Advances in Neural Information Processing Systems 20}, pages 161--168. Curran
  Associates, Inc., 2008.

\bibitem{ByrdNoceSchn1994}
R.~H Byrd, J.~Nocedal, and R.~B Schnabel.
\newblock Representations of quasi \textsc{N}ewton matrices and their use in
  limited memory methods.
\newblock {\em Mathematical Programming}, 63:129--156, 1994.

\bibitem{ByrdNoceHansSing:2015}
R.H. Byrd, S.L. Hansen, J.~Nocedal, and Y.Singer.
\newblock {A Stochastic Quasi-Newton Method for Large-Scale Optimization}.
\newblock Technical Report arXiv:1401.7020, arXiv, 2015.

\bibitem{CartisScheinberg2014}
C.~Cartis and K.~Scheinberg.
\newblock Global convergence rate analysis of unconstrained optimization
  methods based on probabilistic models.
\newblock Technical Report, ISE, Lehigh, 2015.

\bibitem{ChenMenickellyScheinberg2014}
C.~Chen, M.~Menickelly, and K.~Scheinberg.
\newblock Stochastic optimization using a trust-region method and random
  models.
\newblock 2015.

\bibitem{cho2014learning}
Kyunghyun Cho, Bart Van~Merri{\"e}nboer, Caglar Gulcehre, Dzmitry Bahdanau,
  Fethi Bougares, Holger Schwenk, and Yoshua Bengio.
\newblock Learning phrase representations using rnn encoder-decoder for
  statistical machine translation.
\newblock {\em arXiv:1406.1078}, 2014.

\bibitem{ciregan2012multi}
Dan Ciregan, Ueli Meier, and J{\"u}rgen Schmidhuber.
\newblock Multi-column deep neural networks for image classification.
\newblock In {\em Computer Vision and Pattern Recognition (CVPR), 2012 IEEE
  Conference on}, pages 3642--3649. IEEE, 2012.

\bibitem{Curt2016}
Frank Curtis.
\newblock A self-correcting variable-metric algorithm for stochastic
  optimization.
\newblock In {\em Proceedings of The 33rd International Conference on Machine
  Learning}, page 632–641, 2016.

\bibitem{dahl2013improving}
George~E Dahl, Tara~N Sainath, and Geoffrey~E Hinton.
\newblock Improving deep neural networks for lvcsr using rectified linear units
  and dropout.
\newblock In {\em 2013 IEEE International Conference on Acoustics, Speech and
  Signal Processing}, pages 8609--8613. IEEE, 2013.

\bibitem{Bengiononconvex}
Yann~N Dauphin, Razvan Pascanu, Caglar Gulcehre, Kyunghyun Cho, Surya Ganguli,
  and Yoshua Bengio.
\newblock Identifying and attacking the saddle point problem in
  high-dimensional non-convex optimization.
\newblock In {\em Advances in Neural Information Processing Systems}, pages
  2933--2941, 2014.

\bibitem{NIPS2014_5258}
A.~Defazio, F.~Bach, and S.~Lacoste-Julien.
\newblock Saga: A fast incremental gradient method with support for
  non-strongly convex composite objectives.
\newblock In Z.~Ghahramani, M.~Welling, C.~Cortes, N.D. Lawrence, and K.Q.
  Weinberger, editors, {\em Advances in Neural Information Processing Systems
  27}, pages 1646--1654. Curran Associates, Inc., 2014.

\bibitem{deng2014deep}
Li~Deng and Dong Yu.
\newblock Deep learning.
\newblock {\em Signal Processing}, 7:3--4, 2014.

\bibitem{elkordy1993neural}
M.F. Elkordy, K.C. Chang, and G.C. Lee.
\newblock Neural networks trained by analytically simulated damage states.
\newblock {\em Journal of Computing in Civil Engineering}, 7(2):130--145, 1993.

\bibitem{ErdoMont2015}
M.~A. Erdogdu and A.~Montanari.
\newblock Convergence rates of sub-sampled newton methods.
\newblock {\em arXiv:1508.02810}, 2015.

\bibitem{flood1994neural}
I.~Flood and N.~Kartam.
\newblock Neural networks in civil engineering {I}: Principles and
  understanding.
\newblock {\em Journal of computing in civil engineering}, 8(2):131--148, 1994.

\bibitem{doi:10.1137/110830629}
Michael~P. Friedlander and Mark Schmidt.
\newblock Hybrid deterministic-stochastic methods for data fitting.
\newblock {\em SIAM Journal on Scientific Computing}, 34(3):A1380--A1405, 2012.

\bibitem{gent1992predicting}
C.R. Gent and C.P. Sheppard.
\newblock Predicting time series by a fully connected neural network trained by
  back propagation.
\newblock {\em Computing and Control Engineering Journal}, 3(3):109--112, 1992.

\bibitem{ghadimilannonconvex1}
S.~Ghadimi and G.~Lan.
\newblock Stochastic first-and zeroth-order methods for nonconvex stochastic
  programming.
\newblock {\em SIOPT}, 23(4):2341--2368, 2013.

\bibitem{ghadimilannonconvex2}
S.~Ghadimi and G.~Lan.
\newblock Accelerated gradient methods for nonconvex nonlinear and stochastic
  programming.
\newblock {\em Mathematical Programming}, 156(1-2):59--99, 2016.

\bibitem{GneccoSanguineti2008}
G.~Gnecco and M.~Sanguineti.
\newblock Approximation error bounds via rademacher's complexity.
\newblock {\em Applied Mathematical Sciences}, 2(4):153 -- 176, 2008.

\bibitem{gouldgltr}
Nicholas~IM Gould, Stefano Lucidi, Massimo Roma, and Philippe~L Toint.
\newblock Solving the trust-region subproblem using the lanczos method.
\newblock {\em SIAM Journal on Optimization}, 9(2):504--525, 1999.

\bibitem{GoldGoweRich2016}
Robert Gower, Donald Goldfarb, and Peter Richtarik.
\newblock Stochastic block bfgs: Squeezing more curvature out of data.
\newblock In {\em Proceedings of The 33rd International Conference on Machine
  Learning}, page 1869–1878, 2016.

\bibitem{GrattonEtAl2017}
S.~Gratton, C.~W. Royer, L.~N. Vicente, and Z.~Zhang.
\newblock Complexity and global rates of trust-region methods based on
  probabilistic models.
\newblock Technical Report 17-09, Dept. Mathematics, Univ. Coimbra, 2017.

\bibitem{graves2013speech}
Alex Graves, Abdel-rahman Mohamed, and Geoffrey Hinton.
\newblock Speech recognition with deep recurrent neural networks.
\newblock In {\em 2013 IEEE international conference on acoustics, speech and
  signal processing}, pages 6645--6649. IEEE, 2013.

\bibitem{guo2012multi}
Zhenhai Guo, Weigang Zhao, Haiyan Lu, and Jianzhou Wang.
\newblock Multi-step forecasting for wind speed using a modified emd-based
  artificial neural network model.
\newblock {\em Renewable Energy}, 37(1):241--249, 2012.

\bibitem{hagan1996neural}
Martin~T Hagan, Howard~B Demuth, Mark~H Beale, and Orlando De~Jes{\'u}s.
\newblock {\em Neural network design}, volume~20.
\newblock PWS publishing company Boston, 1996.

\bibitem{haykin1994neural}
Simon Haykin.
\newblock Neural networks, a comprehensive foundation.
\newblock 1994.

\bibitem{he2015delving}
Kaiming He, Xiangyu Zhang, Shaoqing Ren, and Jian Sun.
\newblock Delving deep into rectifiers: Surpassing human-level performance on
  imagenet classification.
\newblock In {\em Proceedings of the IEEE International Conference on Computer
  Vision}, pages 1026--1034, 2015.

\bibitem{hinton2012deep}
Geoffrey Hinton, Li~Deng, Dong Yu, George~E Dahl, Abdel-rahman Mohamed, Navdeep
  Jaitly, Andrew Senior, Vincent Vanhoucke, Patrick Nguyen, Tara~N Sainath,
  et~al.
\newblock Deep neural networks for acoustic modeling in speech recognition: The
  shared views of four research groups.
\newblock {\em IEEE Signal Processing Magazine}, 29(6):82--97, 2012.

\bibitem{HochreiterS97}
S.~Hochreiter and J.~Schmidhuber.
\newblock Long short-term memory.
\newblock {\em Neural Computation}, 9:1735--1780, 1997.

\bibitem{hornik1989multilayer}
Kurt Hornik, Maxwell Stinchcombe, and Halbert White.
\newblock Multilayer feedforward networks are universal approximators.
\newblock {\em Neural networks}, 2(5):359--366, 1989.

\bibitem{ji20133d}
Shuiwang Ji, Wei Xu, Ming Yang, and Kai Yu.
\newblock 3d convolutional neural networks for human action recognition.
\newblock {\em IEEE transactions on pattern analysis and machine intelligence},
  35(1):221--231, 2013.

\bibitem{johnson2013accelerating}
R.~Johnson and T.~Zhang.
\newblock Accelerating stochastic gradient descent using predictive variance
  reduction.
\newblock In C.J.C. Burges, L.~Bottou, M.~Welling, Z.~Ghahramani, and K.Q.
  Weinberger, editors, {\em Advances in Neural Information Processing Systems
  26}, pages 315--323. 2013.

\bibitem{karpathy2014large}
Andrej Karpathy, George Toderici, Sanketh Shetty, Thomas Leung, Rahul
  Sukthankar, and Li~Fei-Fei.
\newblock Large-scale video classification with convolutional neural networks.
\newblock In {\em Proceedings of the IEEE conference on Computer Vision and
  Pattern Recognition}, pages 1725--1732, 2014.

\bibitem{kartam1997artificial}
N.~Kartam, I.~Flood, and J.~H Garrett.
\newblock Artificial neural networks for civil engineers: fundamentals and
  applications.
\newblock American Society of Civil Engineers, 1997.

\bibitem{krizhevsky2012imagenet}
Alex Krizhevsky, Ilya Sutskever, and Geoffrey~E Hinton.
\newblock Imagenet classification with deep convolutional neural networks.
\newblock In {\em Advances in neural information processing systems}, pages
  1097--1105, 2012.

\bibitem{Lan2012}
G.~Lan.
\newblock An optimal method for stochastic composite optimization.
\newblock {\em Mathematical Programming}, 133(1):365--397, 2012.

\bibitem{lecun-98b}
Y.~LeCun, L.~Bottou, G.~Orr, and K.~Muller.
\newblock { Efficient BackProp}.
\newblock In {\em Neural Networks: Tricks of the trade}. 1998.

\bibitem{lecun1995convolutional}
Yann LeCun and Yoshua Bengio.
\newblock Convolutional networks for images, speech, and time series.
\newblock {\em The handbook of brain theory and neural networks},
  3361(10):1995, 1995.

\bibitem{lecun2015deep}
Yann LeCun, Yoshua Bengio, and Geoffrey Hinton.
\newblock Deep learning.
\newblock {\em Nature}, 521(7553):436--444, 2015.

\bibitem{lukovsevivcius2009reservoir}
M.~Luko{\v{s}}evi{\v{c}}ius and H.~Jaeger.
\newblock Reservoir computing approaches to recurrent neural network training.
\newblock {\em Computer Science Review}, 3(3):127--149, 2009.

\bibitem{maas2013rectifier}
Andrew~L Maas, Awni~Y Hannun, and Andrew~Y Ng.
\newblock Rectifier nonlinearities improve neural network acoustic models.
\newblock In {\em Proc. ICML}, volume~30, 2013.

\bibitem{martenshfpaper}
J.~Martens.
\newblock Deep learning via hessian-free optimization.
\newblock In {\em Proceedings of the 27th International Conference on Machine
  Learning (ICML-10)}, pages 735--742, 2010.

\bibitem{martenschapter}
J.~Martens and I.~Sutskever.
\newblock Training deep and recurrent networks with hessian-free optimization.
\newblock In {\em Neural Networks: Tricks of the Trade}, pages 479--535.
  Springer, 2012.

\bibitem{deo2001neural}
C.~McDeo, A.~Jha, A.S. Chaphekar, and K.~Ravikant.
\newblock Neural networks for wave forecasting.
\newblock {\em Ocean Engineering}, 28(7):889--898, 2001.

\bibitem{mikolov2010recurrent}
Tomas Mikolov, Martin Karafi{\'a}t, Lukas Burget, Jan Cernock{\`y}, and Sanjeev
  Khudanpur.
\newblock Recurrent neural network based language model.
\newblock In {\em Interspeech}, volume~2, page~3, 2010.

\bibitem{negativecurvatures}
E.~Mizutani and S.~E. Dreyfus.
\newblock Second-order stagewise backpropagation for hessian-matrix analyses
  and investigation of negative curvature.
\newblock {\em Neural Networks}, 21(2):193--203, 2008.

\bibitem{NemJudLanSha2009}
A~Nemirovski, Juditsky, G~Lan, and A~Shapiro.
\newblock {\em SIAM Journal on optimization}, 19.

\bibitem{Nesterov}
Yu. Nesterov.
\newblock {\em Introductory Lectures on Convex Optimization}.
\newblock Kluwer Academic Publsihers, Boston, MA, 2004.

\bibitem{Sarah2017}
L.~M. Nguyen, J.~Liu, K.~Scheinberg, and M.~Tak\'{a}\v{c}.
\newblock Sarah: A novel method for machine learning problems using stochastic
  recursive gradient.
\newblock {\em arXiv:1703.00102}, 2017.

\bibitem{NocedalWright06}
Jorge Nocedal and Stephen~J. Wright.
\newblock {\em Numerical optimization}.
\newblock Springer Series in Operations Research and Financial Engineering.
  Springer, New York, second edition, 2006.

\bibitem{PascanuDGB14}
R.~Pascanu, Y.~N. Dauphin, S.~Ganguli, and Y.~Bengio.
\newblock On the saddle point problem for non-convex optimization.
\newblock {\em CoRR}, abs/1405.4604, 2014.

\bibitem{pasupathyinformstutorial}
R.~Pasupathy and S.~Ghosh.
\newblock Simulation optimization: A concise overview and implementation guide.
\newblock In {\em {TutORials} in Operations Research}, chapter~7, pages
  122--150. INFORMS, 2013.

\bibitem{pearlmutter}
B.~A. Pearlmutter.
\newblock Fast exact multiplication by the hessian.
\newblock {\em Neural computation}, 6(1):147--160, 1994.

\bibitem{Pol64}
B.~T. Polyak.
\newblock Some methods of speeding up the convergence of iteration methods.
\newblock {\em USSR Computational Mathematics and Mathematical Physics},
  4:791--803, 1964.

\bibitem{Robbins:1951ko}
H.~Robbins and S.~Monro.
\newblock {A Stochastic Approximation Method}.
\newblock {\em The Annals of Mathematical Statistics}, 22(3):400--407,
  September 1951.

\bibitem{RoKhoMah2016I}
F.~Roosta-Khorasani and M.~W. Mahoney.
\newblock Sub-sampled newton methods i: Globally convergent algorithms.
\newblock {\em arXiv:1601.047388}, 2016.

\bibitem{RoKhoMah2016II}
F.~Roosta-Khorasani and M.~W. Mahoney.
\newblock Sub-sampled newton methods ii: Local convergence rates.
\newblock {\em arXiv:1601.047388}, 2016.

\bibitem{backprop}
D.~E. Rumelhart, G.~E. Hinton, and R.~J. Williams.
\newblock Neurocomputing: Foundations of research.
\newblock chapter Learning Representations by Back-propagating Errors, pages
  696--699. MIT Press, Cambridge, MA, USA, 1988.

\bibitem{ILSVRC15}
O.~Russakovsky, J.~Deng, H.~Su, J.~Krause, S.~Satheesh, S.~Ma, Zh. Huang,
  A.~Karpathy, A.~Khosla, M.~Bernstein, A.~C. Berg, and L.~Fei-Fei.
\newblock {ImageNet Large Scale Visual Recognition Challenge}.
\newblock {\em International Journal of Computer Vision (IJCV)},
  115(3):211--252, 2015.

\bibitem{RuszShapiroBook}
A.~Ruszczynski and A.~Shapiro, editors.
\newblock {\em Stochastic Programming}.
\newblock Handbooks in Operations Research and Management Science, Volume 10.
  Elsevier, Amsterdam, 2003.

\bibitem{schmidhuber2015deep}
J{\"u}rgen Schmidhuber.
\newblock Deep learning in neural networks: An overview.
\newblock {\em Neural Networks}, 61:85--117, 2015.

\bibitem{schmidt2013minimizing}
M.~Schmidt, N.~LeRoux, and F.~Bach.
\newblock Minimizing finite sums with the stochastic average gradient.
\newblock {\em arXiv preprint arXiv:1309.2388}, 2013.

\bibitem{Schraudolphetal}
N.~N. Schraudolph, J.~Yu, and S.~Gunter.
\newblock A stochastic quasi-newton method for online convex optimization.
\newblock In Marina Meila and Xiaotong Shen, editors, {\em Proceedings of the
  Eleventh International Conference on Artificial Intelligence and Statistics},
  volume~2 of {\em Proceedings of Machine Learning Research}, pages 436--443,
  San Juan, Puerto Rico, 21--24 Mar 2007. PMLR.

\bibitem{shalev2013accelerated}
S.~Shalev-Shwartz and T.~Zhang.
\newblock Accelerated proximal stochastic dual coordinate ascent for
  regularized loss minimization.
\newblock {\em Mathematical Programming}, pages 1--41, 2013.

\bibitem{simonyan2013deep}
Karen Simonyan, Andrea Vedaldi, and Andrew Zisserman.
\newblock Deep inside convolutional networks: Visualising image classification
  models and saliency maps.
\newblock {\em arXiv:1312.6034}, 2013.

\bibitem{SpallBook}
J.C. Spall.
\newblock {\em Introduction to Stochastic Search and Optimization: Estimation,
  Simulation, and Control}.
\newblock Wiley Series in Discrete Mathematics and Optimization. Wiley, 2005.

\bibitem{steihaug}
Trond Steihaug.
\newblock The conjugate gradient method and trust regions in large scale
  optimization.
\newblock {\em SIAM Journal on Numerical Analysis}, 20(3):626--637, 1983.

\bibitem{Sutskeveretal2013}
I.~Sutskever, J.~Martens, G.~Dahl, and G.~Hinton.
\newblock On the importance of initialization and momentum in deep learning.
\newblock In {\em Proceedings of the 30th International Conference on
  International Conference on Machine Learning - Volume 28}, ICML'13, pages
  III--1139--III--1147. JMLR.org, 2013.

\bibitem{thirumalaiah1998river}
Konda Thirumalaiah and MC~Deo.
\newblock River stage forecasting using artificial neural networks.
\newblock {\em Journal of Hydrologic Engineering}, 3(1):26--32, 1998.

\bibitem{toint}
Ph~L Toint.
\newblock Towards an efficient sparsity exploiting newton method for
  minimization.
\newblock {\em Sparse matrices and their uses}, page 1981, 1981.

\bibitem{tsai1999back}
Ching-Piao Tsai and Tsong-Lin Lee.
\newblock Back-propagation neural network in tidal-level forecasting.
\newblock {\em Journal of Waterway, Port, Coastal, and Ocean Engineering},
  125(4):195--202, 1999.

\bibitem{tsomokos2008fully}
Dimitris~I Tsomokos, Sahel Ashhab, and Franco Nori.
\newblock Fully connected network of superconducting qubits in a cavity.
\newblock {\em New Journal of Physics}, 10(11):113020, 2008.

\bibitem{vap95}
V.~N. Vapnik.
\newblock {\em The Nature of Statistical Learning Theory}.
\newblock Springer-Verlag, 1995.

\bibitem{Wrightsurvey}
S.~J. Wright.
\newblock Optimization algorithms for data analysis.
\newblock {\em IAS/Park City Mathematics Series}, to appear.

\bibitem{Xuetal2016}
P.~Xu, J.~Yang, F.~Roosta-Khorasani, Ch. R\'{e}, and M.~W. Mahoney.
\newblock Sub-sampled newton methods with non-uniform sampling.
\newblock In D.~D. Lee, M.~Sugiyama, U.~V. Luxburg, I.~Guyon, and R.~Garnett,
  editors, {\em Advances in Neural Information Processing Systems 29}, pages
  3000--3008. Curran Associates, Inc., 2016.

\bibitem{Alirezathesis}
A.~Yektamaram.
\newblock {\em Optimization Algorithms for Machine Learning Designed for
  Parallel and Distributed Environments}.
\newblock PhD thesis, ISE Department, Lehigh University, Bethlehem, PA, 2017.

\bibitem{yu2011deep}
Dong Yu and Li~Deng.
\newblock Deep learning and its applications to signal and information
  processing.
\newblock {\em IEEE Signal Processing Magazine}, 28(1):145--154, 2011.

\bibitem{RechHardt}
C.~Zhang, S.~Bengio, M.~Hardt, B.~Recht, and O.~Vinyals.
\newblock Understanding deep learning requires rethinking generalization.
\newblock {\em arXiv:1611.03530}, 2016.

\end{thebibliography}
\end{document}